\documentclass{article}

\usepackage[preprint]{neurips_2026}

\usepackage[utf8]{inputenc} 
\usepackage[T1]{fontenc}    
\usepackage{url}            
\usepackage{booktabs}       
\usepackage{amsfonts}       
\usepackage{amsmath}
\usepackage{nicefrac}       
\usepackage{microtype}      
\usepackage{xcolor}         
\usepackage{graphicx}    
\usepackage{multirow}
\usepackage{wrapfig}

\definecolor{grey}{RGB}{150,150,150}

\usepackage{caption}
\usepackage{tcolorbox}
\usepackage{fontawesome5}
\usepackage[]{hyperref}

\usepackage{pifont}
\newcommand{\cmark}{\textcolor{green!60!black}{\ding{51}}}
\newcommand{\xmark}{\textcolor{red}{\ding{55}}}

\begin{document}

\title{ArchEGraph: A Large-Scale Graph Dataset for Geometry–Topology–Physics Aligned Building Energy Modeling}

%

\author{
    Yihui Li$^{1,2}$, \quad
    Yihui Chen$^{1, \ast}$, \quad
    Kaidi Zha$^{1, \ast}$, \quad
    Xiaoyue Yan$^{3, \ast}$, \quad
    Zhexuan Yu$^{1, \ast}$, \\
    \textbf{Shiqi Dai}$^{1, \ast}$, \quad
    \textbf{Jun Xiao}$^{1}$, \quad
    \textbf{Jun Yin}$^{1}$, \quad
    \textbf{Ramon Elias Weber}$^{2, \dagger}$, \quad
    \textbf{Borong Lin}$^{1, \dagger}$ \\
    \noalign{\pagebreak[0]\vspace{0.5em}}
    $^{1}$Tsinghua University \quad $^{2}$UC Berkeley \quad $^{3}$MIT \\
    \noalign{\pagebreak[0]\vspace{0.5em}}
    \small \texttt{\{liyihui23\}@mails.tsinghua.edu.cn, \{ramon\}@berkeley.edu, \{linbr\}@tsinghua.edu.cn}
}

\renewcommand{\thefootnote}{\fnsymbol{footnote}}
\footnotetext[1]{Equal contribution.}
\footnotetext[2]{Corresponding authors.}

\maketitle
\vspace{-30pt}
\begin{center}
    \includegraphics[width=\linewidth]{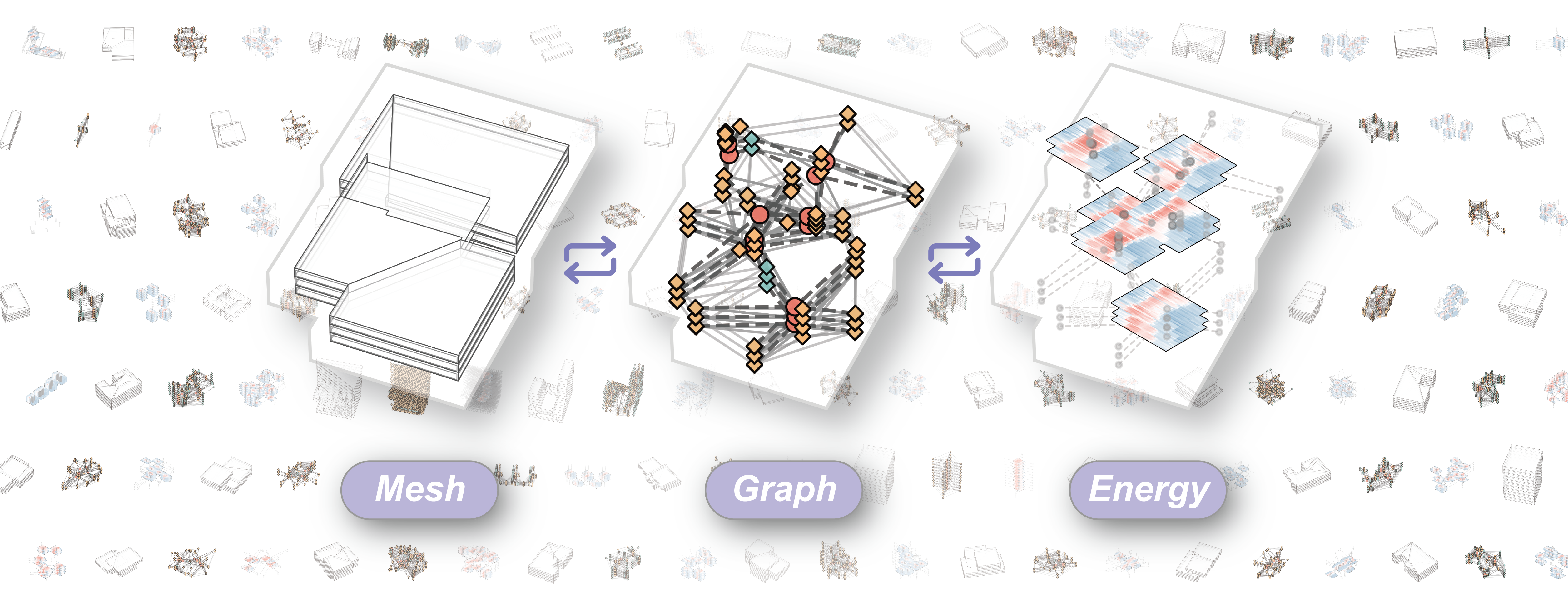}
\end{center}

\newtcbox{\linkbox}{
    colback=gray!5,       
    colframe=gray!20,     
    left=3pt, right=3pt, top=2pt, bottom=2pt,
    boxrule=1pt, arc=10pt, 
    boxsep=2pt,
    on line
}

\begin{center}
    \linkbox{
        \faDatabase \hspace{0.3em} 
        \href{https://huggingface.co/datasets/ArchEGraph/ArchEGraph}{\textsf{Dataset}}
    }
    \hspace{1.5em}
    \linkbox{
        \faGithub \hspace{0.3em} 
        \href{https://github.com/ArchEGraph/ArchEGraph}{\textsf{Code}}
    }
\end{center}

\begin{abstract}

    Accurate estimation of building energy use is essential for achieving carbon neutral and sustainable buildings. To better understand the influence of design decisions on building energy use and calibrate machine learning models that can give architects and engineers rapid design feedback, large-scale datasets are needed that explicitly map building geometry to performance. We present \textbf{ArchEGraph}, a large-scale benchmark dataset that represents buildings as heterogeneous graphs with aligned geometry, topology, weather, and zone-level thermal loads. The dataset contains 5,481 buildings and 49,326 validated building-weather simulation cases. In total, it includes over $1.33 \times 10^5$ space nodes and $1.44 \times 10^6$ face nodes, reflecting substantial geometric and topological complexity. Based on ArchEGraph, we define two benchmark tasks: (i) graph reconstruction from polygonal meshes, aiming to recover topological structure from geometric representations; and (ii) topology-informed load prediction, which leverages graph structure and temporal weather conditions to forecast zone-level response time series. We further introduce standardized evaluation protocols for both tasks and conduct cross-building and cross-climate generalization experiments to assess model robustness. ArchEGraph provides a unified testbed for studying geometry–topology–physics coupling in building energy modeling, enabling the development and evaluation of scalable and generalizable surrogate models.

\end{abstract}

\section{Introduction}
\vspace{-10pt}
Buildings account for a substantial share of global energy consumption and carbon emissions, making accurate and scalable building energy modeling (BEM) that can inform sustainable design choices critical for achieving carbon-neutrality of cities and countries across the globe~\cite{unepb_2023}. Physics-based Building Performance Simulation (BPS) remains the de facto standard for estimating energy demand~\cite{New2012EnergyPlus}, but its use in design exploration and large-scale analysis is constrained by high computational cost~\cite{zhangReviewMachineLearning2021}. A single annual simulation for a complex building requires solving tightly coupled heat transfer, radiative exchange, and dynamic shading processes, all governed by geometry-dependent numerical computation~\cite{clarke2001energy}. As a result, BPS-based workflows remain impractical for iterative design and large-scale scenario evaluation~\cite{Wang2023Building3D,reinhartUrbanBuildingEnergy2016a}.

To mitigate this limitation, machine learning-based surrogate models have been developed as fast approximations of BPS~\cite{Roman2020Metamodels,VonKrannichfeldt2025Hybrid}. However, most existing approaches rely on global feature vectors or sequence-based representations that ignore explicit spatial structure and physical connectivity~\cite{Emami2018,Kelly2015UKDALE,Miller2020BDG2,ASHRAEGEP2020}. This results in two core generalization failures: (i) \textit{geometric shift}, where models fail to extrapolate to unseen architectural layouts due to missing structural inductive bias~\cite{douTransferLearningCrossbuilding2025,ribeiroTransferLearningSeasonal2018}; and (ii) \textit{climatic shift}, where learned mappings degrade under differing weather distributions~\cite{pengClimateadaptiveEnergyForecasting2025}. These issues restrict current surrogate models to narrow design distributions and controlled benchmarks.

We argue that building energy prediction should be formulated as a structured physical learning problem, where energy transfer emerges from interactions between spatial zones and envelope components~\cite{New2012EnergyPlus,ashrae_american_society_of_heating_refrigerating_and_air-conditioning_engineers_ashrae_2017}. This naturally induces a graph representation in which nodes correspond to thermal zones and envelope elements, and edges encode geometric adjacency and physical coupling~\cite{Alymani2023GraphML,Wu2025ResidentialBlocks,weberHypergraphModelShows2024,liGeometryGraphAutomation2026}. Graph Neural Networks provide a principled framework for modeling such structured interactions~\cite{Scarselli2009GNN,Jia2024ThermalGAT}; however, their effectiveness is limited by the absence of large-scale benchmarks that jointly capture geometry, topology, weather forcing, and time-resolved physical responses~\cite{Skeie2025,Emami2023BuildingsBench,Krapf2023PointER}.

To address this gap, we introduce \textbf{ArchEGraph}, a large-scale benchmark for geometry–topology–physics coupled learning in building energy modeling. The dataset contains 5,481 building designs and 49,326 valid simulation instances, with over $1.33 \times 10^5$ spatial nodes and $1.44 \times 10^6$ envelope elements. ArchEGraph supports two tasks: (i) \textit{M2G}, reconstructing building topology (the internal zone configuration that is needed for performance prediction) from geometric inputs; and (ii) \textit{G2E}, predicting zone-level load dynamics under temporal weather conditions. All data is generated via physics-based simulation to ensure consistency across geometry, topology and physics, which is not jointly available in current real-world datasets at scale~\cite{Miller2020BDG2,Krapf2023PointER,Zhu2025GlobalBuildingAtlas}. Together with standardized evaluation protocols and controlled cross-building and cross-climate splits, this establishes a unified benchmark for structured physical learning in buildings.

\section{Related Work}
\vspace{-10pt}
\paragraph{Geometry and graph representations for buildings.}
Buildings are heterogeneous physical systems whose energy behavior depends on both geometric form and spatial connectivity among spaces, envelope elements, openings, and boundary conditions. Existing building information models and simulation formats support building-element representation and physical-parameter storage~\cite{ISO16739,LeonSanchez2022CityGML,New2012EnergyPlus}, while semantic schemas improve interoperability across building domains~\cite{Pauwels2017ifcOWL,Janowicz2021BOT,Balaji2018Brick}. Geometric graph abstractions support layout reasoning and spatial organization~\cite{Engelenburg2025MSD,Nauata2020HouseGAN,Alymani2023GraphML}, BIM/geometry processing~\cite{Gan2022BIMGraph}, and thermal-zone conversion~\cite{Mediavilla2023IFCGraphs,Xiao2024CADBEM}. These developments provide important foundations for representing buildings as structured physical systems, especially when geometry, spatial relations, and boundary elements need to be modeled jointly.

\paragraph{Data-driven surrogates for energy prediction.}
Building performance can be estimated through physics-based simulation or data-driven surrogates. Physics-based simulation provides interpretable performance estimates, while data-driven surrogates enable faster prediction for large-scale design exploration and optimization. Surrogate models based on regressors, tree ensembles, MLPs, recurrent networks, and hybrid approaches have been widely studied for building energy prediction~\cite{Roman2020Metamodels,VonKrannichfeldt2025Hybrid}. More recently, graph neural networks have provided a natural way to model irregular building topologies through message passing~\cite{Scarselli2009GNN}, and have been applied to thermal-load and energy prediction tasks~\cite{Jia2024ThermalGAT,Wu2025ResidentialBlocks}. Together, these studies show the potential of structure-aware learning methods for connecting building geometry, operational context, and energy performance.

\paragraph{Datasets for building geometry and energy modeling.}
Existing datasets only partially support geometry-aware energy learning, as summarized in Table~\ref{tab:building_dataset_comparison}. Recent 3D building geometry and layout datasets provide increasingly large and detailed geometric resources, covering exterior building models, point clouds, meshes, indoor layouts, and floor-plan annotations across different spatial scales~\cite{Peters2022_3DBAG,Zhu2025GlobalBuildingAtlas,Selvaraju2021BuildingNet,Wang2023Building3D,Cruz2021ZInD,Zheng2020Structured3D}. These datasets support reconstruction, semantic labeling, and layout estimation, but are rarely coupled with the physical-response data needed for energy modeling. Conversely, energy/load datasets provide measured or simulated time-series observations for forecasting, disaggregation, and benchmarking~\cite{Emami2023BuildingsBench,Kelly2015UKDALE,Murray2017REFIT,Makonin2016AMPds,PecanStreet2018,Miller2020BDG2,ASHRAEGEP2020}, but are typically geometry-agnostic. Existing geometry-energy resources link the two only partially through reference templates, urban exterior models, static certificate labels, or synthetic operation data~\cite{Deru2011DOEReference,LeonSanchez2022CityGML,Krapf2023PointER,Li2021SyntheticOperation}. This motivates datasets that jointly align 3D geometry, space-surface topology, weather/context variables, and time-resolved energy/load responses.

\vspace{-15pt}

\begin{table}[htbp]
\centering
\caption{Comparison of representative building geometry and energy/load datasets. \textit{3D Geo} denotes building-level geometric representation (e.g., massing or envelope). \textit{Topology} indicates space- or zone-level relationships between interior regions and their boundary surfaces. \textit{Energy/Load} refers to energy or other physical response labels, and \textit{Weather/Context} denotes climate or simulation boundary conditions.}
\scriptsize
\setlength{\tabcolsep}{3.5pt}
\renewcommand{\arraystretch}{0.9}
\begin{tabular}{lcccccc}
\toprule
Dataset 
& Scale 
& 3D Geo 
& Topology
& Energy/Load 
& Weather/Context 
& Temporal Resolution \\
\midrule

\multicolumn{7}{l}{\textit{3D building geometry and layout datasets}} \\
3DBAG~\cite{Peters2022_3DBAG}
& 10M bldgs
& \cmark
& \xmark
& --
& --
& Static\\

GlobalBuildingAtlas~\cite{Zhu2025GlobalBuildingAtlas}
& 2.7B bldgs
& \cmark
& \xmark
& --
& --
& Static\\

BuildingNet~\cite{Selvaraju2021BuildingNet}
& 2k bldgs
& \cmark
& \xmark
& --
& --
& Static\\

Building3D~\cite{Wang2023Building3D}
& 160k bldgs
& \cmark
& \xmark
& --
& --
& Static\\

ZInD~\cite{Cruz2021ZInD}
& 1.5k homes
& \cmark
& \cmark
& --
& --
& Static\\

Structured3D~\cite{Zheng2020Structured3D}
& 3.5K houses
& \cmark
& \cmark
& --
& --
& Static\\

\midrule
\multicolumn{7}{l}{\textit{Building energy and load datasets}} \\
BuildingsBench~\cite{Emami2023BuildingsBench}
& 900k bldgs
& --
& --
& \cmark
& \cmark
& 1h\\

UK-DALE~\cite{Kelly2015UKDALE}
& 5 homes
& --
& --
& \cmark
& \xmark
& 6s\\

REFIT~\cite{Murray2017REFIT}
& 20 homes
& --
& --
& \cmark
& \xmark
& 8s\\

AMPds~\cite{Makonin2016AMPds}
& 1 home
& --
& --
& \cmark
& \cmark
& 1min\\


BDG2~\cite{Miller2020BDG2}
& 1.6k bldgs
& --
& --
& \cmark
& \cmark
& 1h\\

ASHRAE GEP III~\cite{ASHRAEGEP2020}
& 1.5k bldgs
& --
& --
& \cmark
& \cmark
& 1h\\

\midrule
\multicolumn{7}{l}{\textit{Datasets with both building geometry and energy labels}} \\

PointER~\cite{Krapf2023PointER}
& 1M bldgs
& \cmark
& \xmark
& \cmark
& \xmark
& Static\\

DOE Ref. Buildings~\cite{Deru2011DOEReference}
& 16 archetypes
& \cmark
& \xmark
& \cmark
& \cmark
& 1 h\\

AlphaBuilding~\cite{Li2021SyntheticOperation}
& 16 archetypes
& \cmark
& \xmark
& \cmark
& \cmark
& 10min\\


\midrule
\textbf{ArchEGraph (Ours)}
& \textbf{5.5k bldgs \& 49k cases}
& \cmark
& \cmark
& \cmark
& \cmark
& \textbf{1h} \\

\bottomrule
\end{tabular}
\label{tab:building_dataset_comparison}
\end{table}

\section{ArchEGraph dataset}
\vspace{-10pt}
In this section, we first provide an overview of the dataset composition, followed by its formal data schema. Finally, we analyze the statistical characteristics of the dataset to demonstrate its diversity in terms of geometry, climate, and physical response. Detailed sample cases and the comprehensive construction pipeline are provided in Appendix~\ref{si:dataset_details}. 

\subsection{Overview}

\textbf{ArchEGraph} is a graph-centered dataset that represents buildings as heterogeneous graphs, where geometric structures are encoded as graph topology and physical responses are aligned with graph nodes under varying weather conditions. This unified representation enables learning-based modeling of geometry–topology–physics coupling.

The dataset comprises two complementary sources of building geometry: a curated subset (\textbf{ArchEGraph-M}, 580 manually designed cases) and a synthetic subset (\textbf{ArchEGraph-P}, 4,901 parametric-generated cases), resulting in a total of $N=5{,}481$ unique building geometries. By pairing these buildings with up to $M=64$ distinct global weather sequences across climate regions and retaining only quality-controlled outputs, we obtain $K=49,326$ high-fidelity energy simulation cases.

The construction of ArchEGraph involved overcoming three critical technical bottlenecks:
\begin{itemize}
    \item \textbf{Geometric robustness}: Developing an automated pipeline to convert heterogeneous raw meshes into watertight, convex, and graph-compatible representations.
    \item \textbf{Simulation scalability}: Implementing a stable execution framework for large-scale simulations under thousands of geometry-weather combinations.
    \item \textbf{Multi-modal alignment}: Ensuring strict cross-modal consistency to map zone-level simulation outputs (physics) back to the spatial nodes of the graph.
\end{itemize}

\begin{figure}[htbp]
    \centering
    \includegraphics[width=\linewidth]{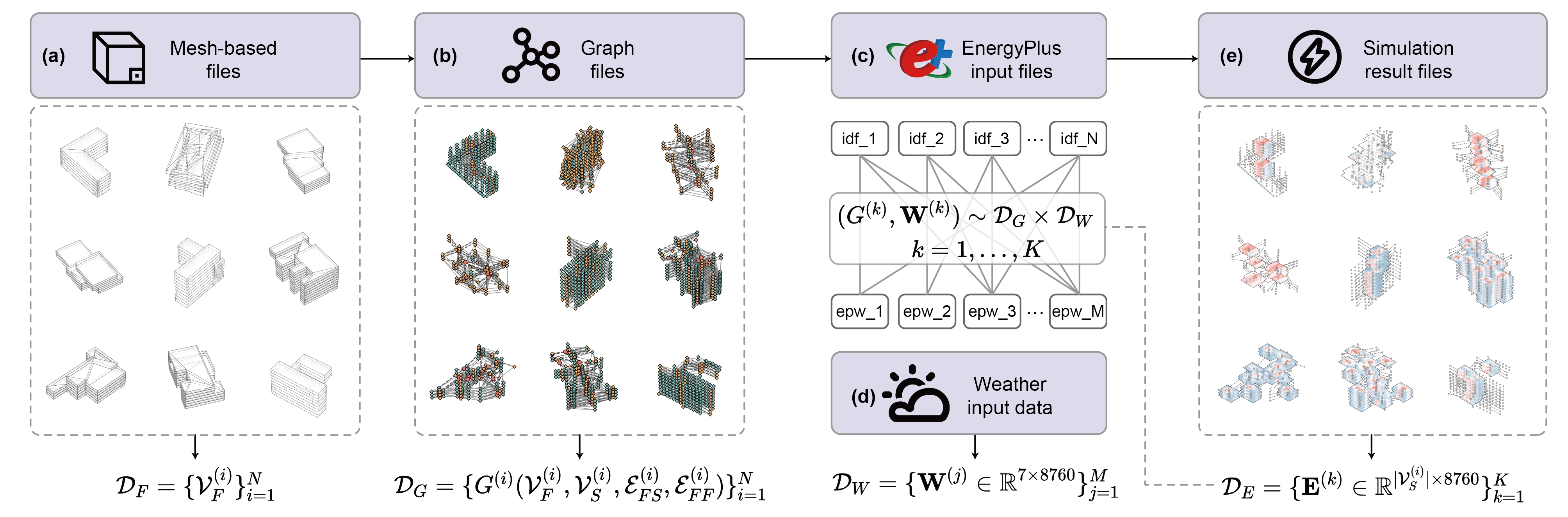}
    \caption{Overview of the ArchEGraph construction workflow. (a) Polygonal mesh files are converted into (b) heterogeneous graph representations, which are then coupled with (c) EnergyPlus input templates and (d) weather files for batch simulation. (e) The resulting aligned modalities form $\mathcal{D}_F$ (geometry), $\mathcal{D}_G$ (graph structure), $\mathcal{D}_W$ (weather), and $\mathcal{D}_E$ (zone-level simulation responses). Each energy simulation instance corresponds to a paired weather–building configuration $(G^{(k)}, \mathbf{W}^{(k)})$.}
    \label{fig:pipeline}
\end{figure}

Figure~\ref{fig:pipeline} illustrates the overall data construction process. Building geometries are first converted into structured graph representations, which are then coupled with weather inputs and processed through energy simulation to generate zone-level physical responses. More detailed data generation procedures are described in the Appendix~\ref{si:data_pipeline}.

\vspace{-10pt}

\subsection{Data schema}

ArchEGraph adopts a graph-centric abstraction in which the building topology serves as the structural backbone for aligning geometry, weather conditions, and energy responses. Building upon prior graph-based building representations~\cite{liGeometryGraphAutomation2026,yuPhysicalEmbeddingBuilding2026}, we further extend the framework to support weather-aware analysis while preserving the original geometric--spatial abstraction.

\paragraph{Building representation ($\mathcal{D}_G$)}
The building dataset is represented as a collection of heterogeneous graphs, denoted by $\mathcal{D}_G=\{G^{(i)}\}_{i=1}^N$. Each graph $G^{(i)}=(\mathcal{V}_F^{(i)},\mathcal{V}_S^{(i)},\mathcal{E}_{FF}^{(i)},\mathcal{E}_{FS}^{(i)})$ consists of face-level nodes $\mathcal{V}_F$, spatial nodes $\mathcal{V}_S$, face--face relations $\mathcal{E}_{FF}$, and face--space relations $\mathcal{E}_{FS}$. This representation jointly encodes geometric decomposition and spatial semantics, while remaining compatible with weather-conditioned physical modeling.

\paragraph{Geometry source ($\mathcal{D}_F$)}
Each graph is derived from a polygonal mesh list stored in the geometry dataset, forming the geometric source representation $\mathcal{D}_F = \{\mathcal{V}_F^{(i)}\}_{i=1}^N$, where the segmented mesh faces $\mathcal{V}_F$ provide the raw geometric basis from which graph nodes and connectivity are constructed.

\paragraph{Weather conditions ($\mathcal{D}_W$)}
We incorporate $M=64$ global hourly weather sequences $\mathcal{D}_W = \{\mathbf{W}^{(j)} \in \mathbb{R}^{7 \times 8760}\}_{j=1}^M$, each sequence includes seven critical climatic variables (e.g., dry-bulb temperature, global horizontal radiation, wind speed) that drive the thermal dynamics of the buildings.

\paragraph{Energy responses ($\mathcal{D}_E$)}
The energy dataset aligns simulation outputs with the spatial nodes of the graph. For each building-weather pair $(G^{(k)}, \mathbf{W}^{(k)})$, we record hourly zone-level load responses in $\mathcal{D}_E = \{\mathbf{E}^{(k)} \in \mathbb{R}^{|\mathcal{V}_S^{(i)}| \times 8760}\}_{k=1}^K$.

This alignment enables supervised learning on graphs, where the model learns to predict the temporal evolution of physical states based on static graph topology and dynamic external forcing.

\subsection{Statistical characteristics}

We performed a comprehensive statistical analysis focusing on three dimensions: geometric \& topological complexity, climatic coverage, and physical response variance. Detailed results can be found in Appendix~\ref{si:stat}.

\paragraph{Geometric complexity}

\begin{wraptable}{r}{0.55\textwidth} 
    \centering
    \vspace{-0pt} 
    \caption{Summary of ArchEGraph scale and metrics.}
    \label{tab:core_stats_transposed}
    \small 
    \begin{tabular}{lrrr}
    \toprule
    \textbf{Graph element} & \textbf{P} & \textbf{M} & \textbf{All(P+M)} \\
    \midrule
    Face nodes$|\mathcal{V}_F|$  
    & 1,242,324 & 203,853 & 1,446,177 \\
    Space nodes$|\mathcal{V}_S|$
    & 119,265 & 14,209 & 133,474 \\
    FF edges$|\mathcal{E}_{FF}|$    
    & 3,846,552 & 553,239 & 4,399,791 \\
    FS edges$|\mathcal{E}_{FS}|$        
    & 1,422,826 & 140,943 & 1,563,769 \\    
    \bottomrule
    \end{tabular}
    \vspace{-1em} 
\end{wraptable}

\begin{figure}[htbp]
    \centering
    \includegraphics[width=\linewidth]{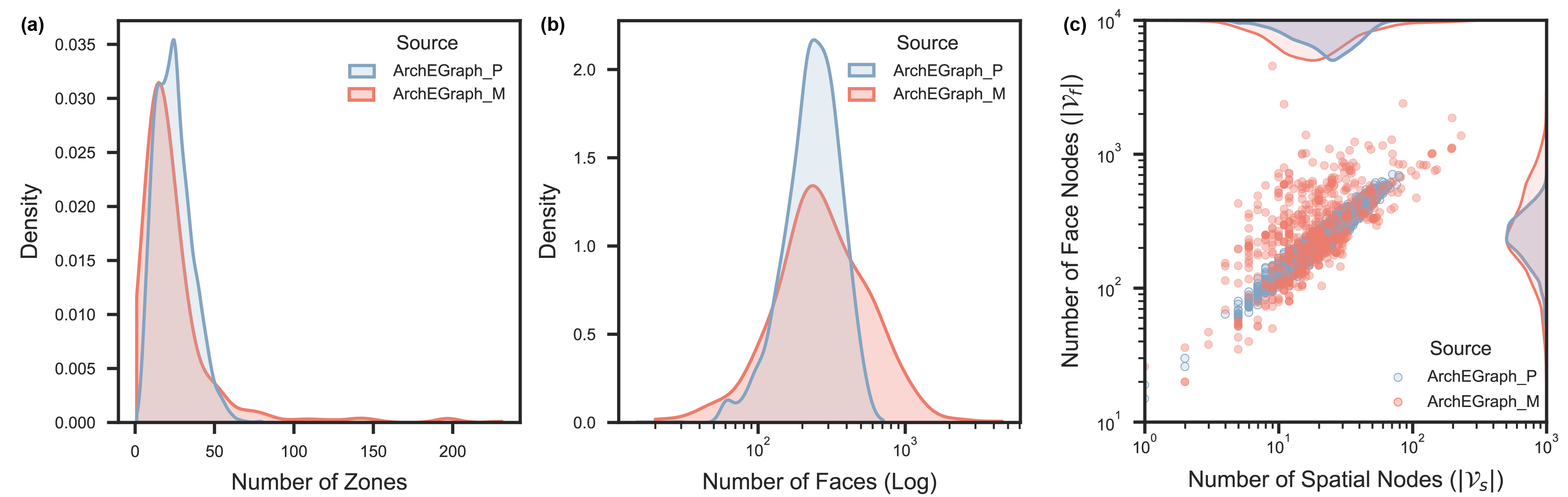}
    \caption{Cross-split geometric statistics of ArchEGraph-P and ArchEGraph-M. (a) Kernel density of the number of spatial nodes (zones) per building. (b) Kernel density of face-node counts (log-scaled axis). (c) Joint distribution between spatial-node count $|\mathcal{V}_S|$ and face-node count $|\mathcal{V}_F|$ with marginal densities, showing scale coupling and diversity across the two subsets.}
    \label{fig:geometry_stat}
\end{figure}

We evaluate the structural diversity of ArchEGraph through the lens of graph scale, connectivity, and hierarchical organization. As summarized in Table~\ref{tab:core_stats_transposed}, the dataset encompasses over 1.4M face nodes and 133k spatial nodes, establishing a high-density representation of architectural environments. The subset ArchEGraph-P provides dense coverage of parametric variations, while ArchEGraph-M introduces long-tail instances with complex, irregular topologies (see Figure~\ref{fig:geometry_stat}). 

Building scales span nearly three orders of magnitude, with average $|\mathcal{V}_F|=263.9$ and $|\mathcal{V}_S|=24.5$. The joint distribution in Figure~\ref{fig:geometry_stat}c reveals a strong coupling between spatial and face-node counts alongside substantial variance. This multi-modal distribution challenges models to generalize across varying mesh resolutions and spatial partitions, ensuring robustness to both synthetic regularities and real-world architectural idiosyncrasies; additional discussion is provided in Appendix~\ref{si:geo_stat}.

\paragraph{Climatic coverage}

\begin{figure*}[htbp]
    \centering

    \begin{minipage}[b]{0.49\textwidth}
        \centering
        \includegraphics[width=\textwidth]{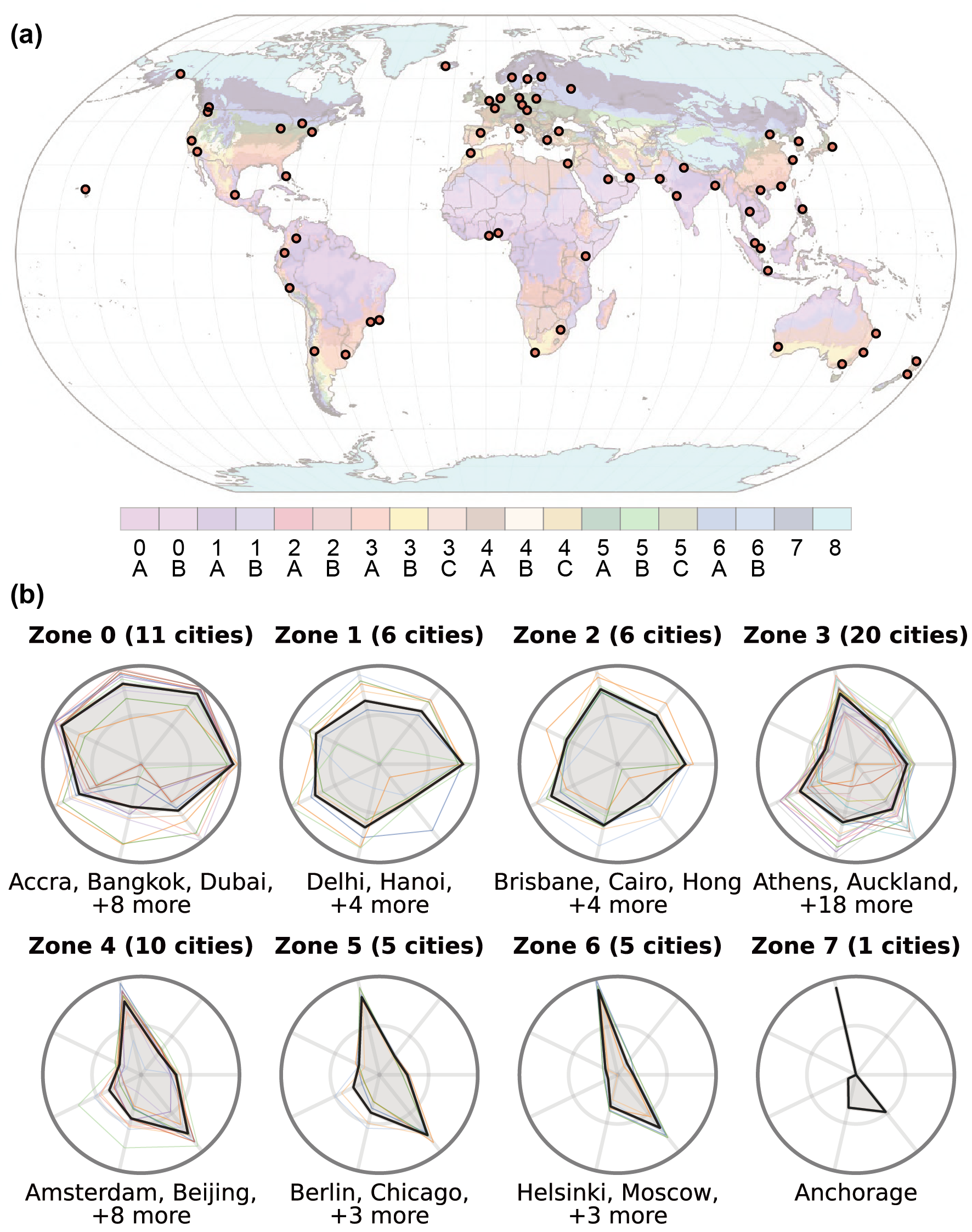}
        \captionof{figure}{Weather profiles across 64 global cities in the dataset. (a) Geographic location of the 64 cities in the ASHRAE climate region map~\cite{ashrae_climate_2021}. (b) Radar map of 7 weather attributes clustering across climate regions.}
        \label{fig:weather_radar}
    \end{minipage}
    \hfill
    \begin{minipage}[b]{0.49\textwidth}
        \centering
        \includegraphics[width=\textwidth]{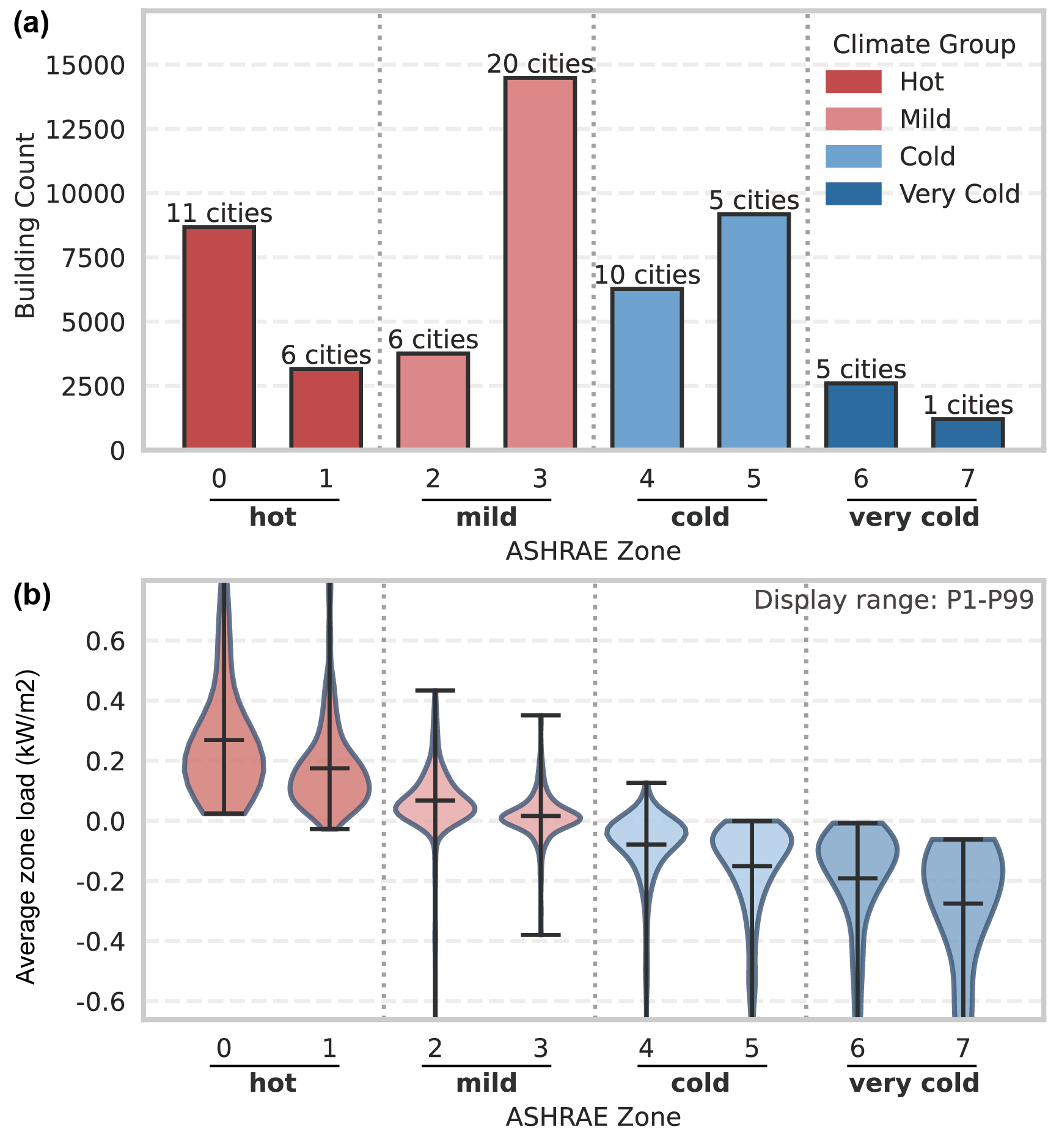}
        \caption{Building samples and energy load across climate regions. (a) Distribution of building samples, showing the total building count and the number of unique cities represented in each climate region. (b) Average zone load intensity, illustrating the distribution of average zone loads ($W/m^2$) across climate classifications ranging from hot to very cold.}
        \label{fig:dataset_energy_anual_load}
    \end{minipage}

\end{figure*}

The dataset exhibits substantial weather diversity supported by both multidimensional meteorological variation and globally distributed sampling. As shown in Figure~\ref{fig:weather_radar}, we characterize 64 representative cities using seven standard meteorological variables widely adopted in building energy simulation, including dry-bulb temperature (db), dew point, relative humidity (rh), solar radiation (measured as Global Horizontal Irradiance (ghi), Direct Normal Irradiance (dni) and Diffuse Horizontal Irradiance (dhi)), and wind speed, which together capture the primary climatic drivers of building energy demand~\cite{Bhandari2012Weather}. The radar profiles in Figure~\ref{fig:weather_radar}b reveal clear and interpretable climate patterns, where tropical regions are dominated by high humidity and stable temperatures, arid regions receive strong solar radiation and low moisture, temperate zones present balanced conditions, and high-latitude cities show reduced solar input and greater variability. In parallel, the geographic distribution spans all inhabited continents and covers cities within a wide latitude range , ensuring balanced representation across major climate regions. Together, these characteristics establish the dataset as a comprehensive and diverse benchmark for evaluating machine learning models under realistic and heterogeneous global weather conditions.

\paragraph{Physical response variance}

The simulated building energy responses exhibit substantial variability across time and space, associated with coupled interactions between climate conditions, building geometry, and dynamic thermal processes. Based on EnergyPlus simulations across 64 cities covering all ASHRAE climate regions~\cite{ashrae_climate_2021} (shown in Figure~\ref{fig:dataset_energy_anual_load}a), the results reveal distinct differences in energy response profiles in terms of magnitude and temporal characteristics, highlighting the joint influence of external weather and intrinsic building properties.

In Figure~\ref{fig:dataset_energy_anual_load}b, the distribution of annual average building loads, defined here as external thermal load, exhibits clear climate-dependent patterns; lower values correspond to higher outward heat transfer. The systematically varying load levels and variances across climate regions indicate consistent coupling between climatic conditions and simulated thermal responses. More details are provided in Appendix~\ref{si:energy_stat}.

Overall, these results suggest that the dataset captures physically consistent climate–energy relationships while preserving substantial intra-class variability. Such properties provide a reliable basis for evaluating geometry-aware and climate-generalizable energy prediction models.

\begin{wrapfigure}[18]{r}{0.48\textwidth}
    \centering
    \vspace{-12pt}
    \includegraphics[width=\linewidth]{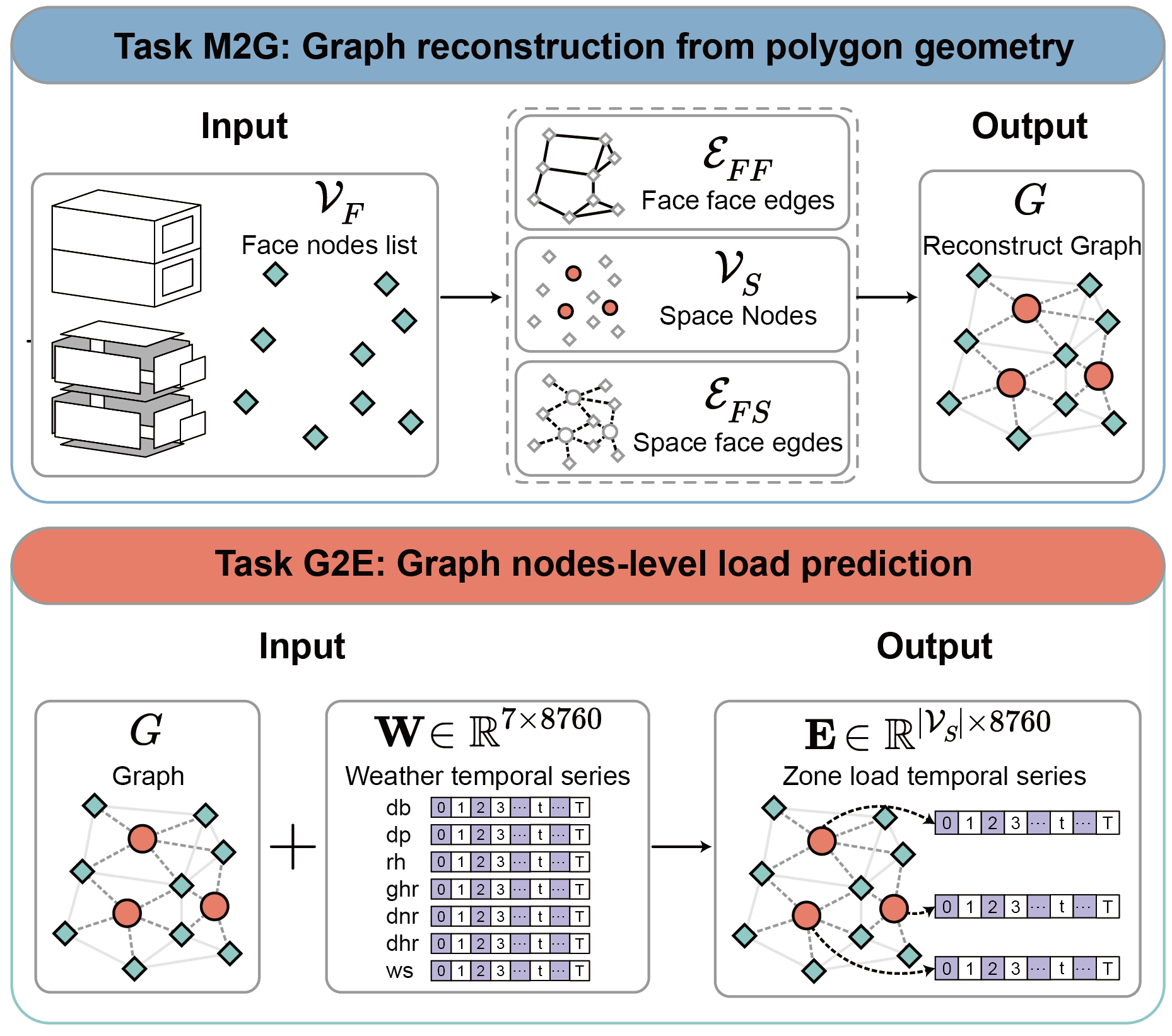} 
    \caption{Task definitions in ArchEGraph. Top: M2G Task, predicting $\mathcal{D}_G$ from polygon face inputs $\mathcal{D}_F$; Bottom: G2E Task, predicting zone-level hourly load $\mathcal{D}_E$ by using graph $\mathcal{D}_G$ and weather condition $\mathcal{D}_W$}
    \label{fig:task}
\end{wrapfigure}

\section{Tasks}
\vspace{-10pt}

We define two tasks on the proposed aligned dataset (Figure~\ref{fig:task}): Mesh-to-Graph reconstruction (M2G) and Graph-to-Energy prediction (G2E).

\subsection{Mesh-to-Graph reconstruction (M2G)}

We formulate Mesh-to-Graph reconstruction as a structured prediction task that infers a heterogeneous building graph from geometric primitives. Given a building composed of a set of faces $\mathcal{V}_F$, each face $f \in \mathcal{V}_F^{(i)}$ is represented by geometric features $(c, R, s)$, where $c \in \mathbb{R}^3$ denotes the centroid, $R \in \mathbb{R}^{3 \times 3}$ denotes the rotation matrix, and $s \in \mathbb{R}^3$ denotes the oriented bounding box size. No topological or semantic information is provided as input. We aim to learn a function:
\begin{equation}
    f_\theta: \mathcal{V}_F^{(i)} \rightarrow G^{(i)} = (\mathcal{V}_S^{(i)}, \mathcal{V}_F^{(i)}, \mathcal{E}_{FF}^{(i)}, \mathcal{E}_{FS}^{(i)}),
\end{equation}
where $\mathcal{V}_S^{(i)}$ denotes space nodes, $\mathcal{E}_{FF}^{(i)}$ represents face–face adjacency relations, and $\mathcal{E}_{FS}^{(i)}$ encodes face–space incidence relations. Supervision is provided by ground-truth building decompositions, where each face is assigned to a unique space, and both $\mathcal{E}_{FF}^{(i)}$ and $\mathcal{E}_{FS}^{(i)}$ are derived from annotated topological relationships.

\subsection{Graph-to-Energy prediction (G2E)}

We formulate Graph-to-Energy prediction as a supervised spatio-temporal learning task over the aligned dataset ${\mathcal{D}_G, \mathcal{D}_W, \mathcal{D}_E}$. Given a building graph $G^{(k)} \in \mathcal{D}_G$ and its corresponding weather sequence $\mathbf{W}^{(k)} \in \mathcal{D}_W$, the objective is to learn a mapping function:
\begin{equation}
    f_\theta: (G^{(k)}, \mathbf{W}^{(k)}) \rightarrow \mathbf{E}^{(k)},
\end{equation}
where $\mathbf{E}^{(k)} \in \mathbb{R}^{|\mathcal{V}_S^{(k)}| \times 8760}$ denotes the zone-level hourly load responses aligned with spatial nodes in the graph. This formulation explicitly models the coupling between heterogeneous graph structure, external climatic forcing, and temporal physical responses, enabling supervised learning of building energy dynamics under varying environmental conditions.

\section{Evaluation}
\vspace{-10pt}
We evaluate the proposed dataset along three dimensions: (i) task-wise performance on structured reconstruction and prediction, and (ii) generalization under cross-building and cross-climate distribution shifts. For more details, see Appendix~\ref{si:eval}.

\subsection{Baselines}
We define the models and evaluation metrics for both benchmark tasks here; detailed full model descriptions and metric definitions are provided in Appendix~\ref{si:metric_and_model_details}.

\paragraph{M2G Baselines}
For M2G, we group baselines into two categories aligned with the topology reconstruction objective:
(1) \textbf{Neural set baselines}: MLP, which applies pointwise linear transformations with no token interaction, and DeepSets~\cite{Zaheer2017DeepSets}, which aggregates set-level context via a sum-pooling encoder;
(2) \textbf{Attention-based set baselines}: Perceiver~\cite{Jaegle2021Perceiver} and SetTransformer~\cite{Lee2019SetTransformer}, which leverage cross-attention or inducing-point attention to model set interactions, and TopoTransformer, which applies a standard Transformer encoder directly over face tokens.
All five models share identical prediction heads for topology reconstruction and differ only in the face encoder design.
We evaluate all methods on three topology components, namely face-face adjacency ($\mathcal{E}_{FF}$), space-node reconstruction ($\mathcal{V}_{S}$), and space-face relations ($\mathcal{E}_{FS}$), using F1 score and accuracy (Acc).

\paragraph{G2E Baselines}
For G2E, we organize baselines into three categories:
(1) \textbf{Non-graph baseline}: WeatherMLP, which predicts node-level energy demand from weather sequences only.
(2) \textbf{Hetero-graph message passing baselines}: F2S and F2S-Attr, both of which perform face-to-space message passing with separate transparent and opaque message branches during spatial propagation. Compared with F2S, F2S-Attr uses stronger normalized MLP blocks and additionally applies attention-weighted message scaling in the spatial propagation stage.
(3) \textbf{Spatial-operator variants}: F2S-GATv2 \cite{Brody2022HowAttentive}, F2S-TransConv \cite{Shi2021MaskedLabelPrediction}, and F2S-GPS \cite{Rampasek2022GraphGPS}, which share the same overall graph-to-energy pipeline and mainly differ in the spatial operator used to update space representations. For a controlled comparison, the spatial-operator variants are implemented with the same input modalities, weather encoder, and temporal GRU decoder. F2S and F2S-Attr follow the same high-level spatial-then-temporal prediction paradigm, while using their own dedicated message-passing parameterizations. We report node-level MAE, MSE, and RMSE.

\subsection{Main results}

\paragraph{M2G Results: Graph Reconstruction}
We evaluate the ability of different methods to reconstruct heterogeneous building graphs from geometric inputs. The results demonstrate how effectively each method captures geometric and topological relationships.

\begin{table}[htbp]
\centering
\scriptsize
\setlength{\tabcolsep}{2pt}
\caption{Graph reconstruction performance results on ArchEGraph-P and ArchEGraph-M.}
\begin{tabular}{l cc cc cc}
\toprule

\multicolumn{7}{c}{\textbf{ArchEGraph-P}} \\
\midrule
Method 
& $\mathcal{E}_{FF}$ (F1) & $\mathcal{E}_{FF}$ (Acc)
& $\mathcal{V}_S$ (F1)    & $\mathcal{V}_S$ (Acc)
& $\mathcal{E}_{FS}$ (F1) & $\mathcal{E}_{FS}$ (Acc) \\
\midrule

MLP 
& $0.9711 \,\textcolor{gray}{\scriptscriptstyle \pm 0.0003}$ 
& $0.9703 \,\textcolor{gray}{\scriptscriptstyle \pm 0.0003}$ 
& $0.9263 \,\textcolor{gray}{\scriptscriptstyle \pm 0.0067}$ 
& $0.9501 \,\textcolor{gray}{\scriptscriptstyle \pm 0.0039}$ 
& $0.3442 \,\textcolor{gray}{\scriptscriptstyle \pm 0.0119}$ 
& $0.8385 \,\textcolor{gray}{\scriptscriptstyle \pm 0.0062}$ \\

DeepSets
& $0.9726 \,\textcolor{gray}{\scriptscriptstyle \pm 0.0008}$ 
& $0.9719 \,\textcolor{gray}{\scriptscriptstyle \pm 0.0008}$ 
& $0.9282 \,\textcolor{gray}{\scriptscriptstyle \pm 0.0057}$ 
& $0.9520 \,\textcolor{gray}{\scriptscriptstyle \pm 0.0028}$ 
& $0.3648 \,\textcolor{gray}{\scriptscriptstyle \pm 0.0086}$ 
& $0.8410 \,\textcolor{gray}{\scriptscriptstyle \pm 0.0053}$ \\

Perceiver 
& $0.9665 \,\textcolor{gray}{\scriptscriptstyle \pm 0.0099}$ 
& $0.9654 \,\textcolor{gray}{\scriptscriptstyle \pm 0.0107}$ 
& $0.9049 \,\textcolor{gray}{\scriptscriptstyle \pm 0.0193}$ 
& $0.9335 \,\textcolor{gray}{\scriptscriptstyle \pm 0.0135}$ 
& $0.3042 \,\textcolor{gray}{\scriptscriptstyle \pm 0.0262}$ 
& $0.8429 \,\textcolor{gray}{\scriptscriptstyle \pm 0.0192}$ \\

SetTransformer 
& $\textbf{0.9758} \,\textcolor{gray}{\scriptscriptstyle \pm 0.0007}$ 
& $\textbf{0.9753} \,\textcolor{gray}{\scriptscriptstyle \pm 0.0007}$ 
& $0.9422 \,\textcolor{gray}{\scriptscriptstyle \pm 0.0007}$ 
& $0.9599 \,\textcolor{gray}{\scriptscriptstyle \pm 0.0007}$ 
& $\textbf{0.4317} \,\textcolor{gray}{\scriptscriptstyle \pm 0.0035}$ 
& $\textbf{0.8810} \,\textcolor{gray}{\scriptscriptstyle \pm 0.0021}$ \\

TopoTransformer 
& $0.9754 \,\textcolor{gray}{\scriptscriptstyle \pm 0.0005}$ 
& $0.9748 \,\textcolor{gray}{\scriptscriptstyle \pm 0.0005}$ 
& $\textbf{0.9443} \,\textcolor{gray}{\scriptscriptstyle \pm 0.0140}$ 
& $\textbf{0.9616} \,\textcolor{gray}{\scriptscriptstyle \pm 0.0110}$ 
& $0.4308 \,\textcolor{gray}{\scriptscriptstyle \pm 0.0022}$ 
& $0.8796 \,\textcolor{gray}{\scriptscriptstyle \pm 0.0020}$ \\

\midrule

\multicolumn{7}{c}{\textbf{ArchEGraph-M}} \\
\midrule
Method 
& $\mathcal{E}_{FF}$ (F1) & $\mathcal{E}_{FF}$ (Acc)
& $\mathcal{V}_S$ (F1)    & $\mathcal{V}_S$ (Acc)
& $\mathcal{E}_{FS}$ (F1) & $\mathcal{E}_{FS}$ (Acc) \\
\midrule

MLP 
& $0.8750 \,\textcolor{gray}{\scriptscriptstyle \pm 0.0116}$ 
& $0.8742 \,\textcolor{gray}{\scriptscriptstyle \pm 0.0106}$ 
& $0.7956 \,\textcolor{gray}{\scriptscriptstyle \pm 0.0166}$ 
& $0.8795 \,\textcolor{gray}{\scriptscriptstyle \pm 0.0087}$ 
& $0.1955 \,\textcolor{gray}{\scriptscriptstyle \pm 0.0043}$ 
& $0.8363 \,\textcolor{gray}{\scriptscriptstyle \pm 0.0167}$ \\

DeepSets 
& $0.8930 \,\textcolor{gray}{\scriptscriptstyle \pm 0.0083}$ 
& $0.8882 \,\textcolor{gray}{\scriptscriptstyle \pm 0.0037}$ 
& $0.8012 \,\textcolor{gray}{\scriptscriptstyle \pm 0.0169}$ 
& $0.8886 \,\textcolor{gray}{\scriptscriptstyle \pm 0.0186}$ 
& $0.2200 \,\textcolor{gray}{\scriptscriptstyle \pm 0.0099}$ 
& $0.8404 \,\textcolor{gray}{\scriptscriptstyle \pm 0.0164}$ \\

Perceiver 
& $0.9009 \,\textcolor{gray}{\scriptscriptstyle \pm 0.0027}$ 
& $0.8827 \,\textcolor{gray}{\scriptscriptstyle \pm 0.0043}$ 
& $0.7533 \,\textcolor{gray}{\scriptscriptstyle \pm 0.0101}$ 
& $0.8599 \,\textcolor{gray}{\scriptscriptstyle \pm 0.0072}$ 
& $0.1883 \,\textcolor{gray}{\scriptscriptstyle \pm 0.0018}$ 
& $0.8467 \,\textcolor{gray}{\scriptscriptstyle \pm 0.0084}$ \\

SetTransformer 
& $0.9006 \,\textcolor{gray}{\scriptscriptstyle \pm 0.0086}$ 
& $\bf{0.8951} \,\textcolor{gray}{\scriptscriptstyle \pm 0.0115}$ 
& $\bf{0.8018} \,\textcolor{gray}{\scriptscriptstyle \pm 0.0126}$ 
& $\bf{0.8970} \,\textcolor{gray}{\scriptscriptstyle \pm 0.0107}$ 
& $\bf{0.2508} \,\textcolor{gray}{\scriptscriptstyle \pm 0.0007}$ 
& $\bf{0.8589} \,\textcolor{gray}{\scriptscriptstyle \pm 0.0019}$ \\

TopoTransformer 
& $\bf{0.9015} \,\textcolor{gray}{\scriptscriptstyle \pm 0.0087}$ 
& $0.8949 \,\textcolor{gray}{\scriptscriptstyle \pm 0.0114}$ 
& $0.7829 \,\textcolor{gray}{\scriptscriptstyle \pm 0.0072}$ 
& $0.8912 \,\textcolor{gray}{\scriptscriptstyle \pm 0.0036}$ 
& $0.2360 \,\textcolor{gray}{\scriptscriptstyle \pm 0.0089}$ 
& $0.8578 \,\textcolor{gray}{\scriptscriptstyle \pm 0.0007}$ \\

\bottomrule
\end{tabular}
\label{tab:graph_construction}
\end{table}

Results in Table~\ref{tab:graph_construction} show three consistent trends. First, ArchEGraph-P consistently outperforms ArchEGraph-M across all metrics, indicating a clear distribution shift from parametric to manually designed geometries. Second, task difficulty is ordered as $\mathcal{E}_{FF}$ (easiest), $\mathcal{V}_S$ (intermediate), and $\mathcal{E}_{FS}$ (hardest), with the gap most pronounced on ArchEGraph-M. Third, \textbf{SetTransformer} and \textbf{TopoTransformer} achieve the best overall performance; \textbf{SetTransformer} is more robust on ArchEGraph-M, while \textbf{TopoTransformer} remains competitive across components.

The difficulty of $\mathcal{E}_{FS}$ stems from its combinatorial nature: correctly assigning each face to a unique space requires resolving global partition consistency across all face–space pairs simultaneously, whereas $\mathcal{E}_{FF}$ and $\mathcal{V}_S$ are comparatively local or count-level predictions. In irregular manually designed geometries (ArchEGraph-M), ambiguous boundary faces shared between zones further increase the error rate, as local geometric features alone are insufficient to determine zone membership without global context. 

\paragraph{G2E Results: Energy Prediction}

We evaluate spatial-temporal energy prediction performance on reconstructed building graphs.

\begin{table}[htbp]
\centering
\scriptsize
\setlength{\tabcolsep}{2pt}
\caption{Energy prediction performance on ArchEGraph-P and ArchEGraph-M datasets.}
\begin{tabular}{lcccccc}
\toprule

Method 
& \multicolumn{3}{c}{\textbf{ArchEGraph-P}}
& \multicolumn{3}{c}{\textbf{ArchEGraph-M}} \\
\cmidrule(lr){2-4} \cmidrule(lr){5-7}
& MAE & MSE & RMSE
& MAE & MSE & RMSE \\
& ( W/m$^2$) & (W/m$^2$) & (W/m$^2$)
& (W/m$^2$) & (W/m$^2$) & (W/m$^2$) \\
\midrule

WeatherMLP        
& $8.845\textcolor{gray}{\scriptscriptstyle \pm 0.548}$ 
& $4.481\textcolor{gray}{\scriptscriptstyle \pm 0.717}$ 
& $6.694\textcolor{gray}{\scriptscriptstyle \pm 0.884}$ 
& $7.540\textcolor{gray}{\scriptscriptstyle \pm 0.867}$ 
& $2.257\textcolor{gray}{\scriptscriptstyle \pm 0.963}$ 
& $4.750\textcolor{gray}{\scriptscriptstyle \pm 1.460}$ \\

F2S               
& $5.781\textcolor{gray}{\scriptscriptstyle \pm 0.332}$ 
& $3.910\textcolor{gray}{\scriptscriptstyle \pm 0.108}$ 
& $6.253\textcolor{gray}{\scriptscriptstyle \pm 0.781}$ 
& $5.159\textcolor{gray}{\scriptscriptstyle \pm 0.365}$ 
& $2.038\textcolor{gray}{\scriptscriptstyle \pm 0.122}$ 
& $4.515\textcolor{gray}{\scriptscriptstyle \pm 0.811}$ \\

F2S-Attr           
& $3.965\textcolor{gray}{\scriptscriptstyle \pm 0.138}$ 
& $1.289\textcolor{gray}{\scriptscriptstyle \pm 0.043}$ 
& $3.590\textcolor{gray}{\scriptscriptstyle \pm 0.610}$ 
& $4.440\textcolor{gray}{\scriptscriptstyle \pm 0.360}$ 
& $1.926\textcolor{gray}{\scriptscriptstyle \pm 0.056}$ 
& $4.389\textcolor{gray}{\scriptscriptstyle \pm 0.628}$ \\

F2S-GATv2         
& $4.331\textcolor{gray}{\scriptscriptstyle \pm 0.221}$ 
& $1.369\textcolor{gray}{\scriptscriptstyle \pm 0.052}$ 
& $3.699\textcolor{gray}{\scriptscriptstyle \pm 0.337}$  
& $4.761\textcolor{gray}{\scriptscriptstyle \pm 0.274}$ 
& $1.941\textcolor{gray}{\scriptscriptstyle \pm 0.061}$ 
& $4.405\textcolor{gray}{\scriptscriptstyle \pm 0.298}$ \\

F2S-TransConv   
& $\mathbf{3.124}\textcolor{gray}{\scriptscriptstyle \pm 0.148}$ 
& $\mathbf{1.100}\textcolor{gray}{\scriptscriptstyle \pm 0.010}$ 
& $\mathbf{3.317}\textcolor{gray}{\scriptscriptstyle \pm 0.154}$ 
& $4.465\textcolor{gray}{\scriptscriptstyle \pm 0.181}$ 
& $1.958\textcolor{gray}{\scriptscriptstyle \pm 0.015}$ 
& $4.425\textcolor{gray}{\scriptscriptstyle \pm 0.165}$ \\

F2S-GPS          
& $3.290\textcolor{gray}{\scriptscriptstyle \pm 0.163}$ 
& $1.141\textcolor{gray}{\scriptscriptstyle \pm 0.018}$ 
& $3.378\textcolor{gray}{\scriptscriptstyle \pm 0.192}$ 
& $\mathbf{3.425}\textcolor{gray}{\scriptscriptstyle \pm 0.147}$ 
& $\mathbf{1.687}\textcolor{gray}{\scriptscriptstyle \pm 0.026}$ 
& $\mathbf{4.108}\textcolor{gray}{\scriptscriptstyle \pm 0.173}$ \\

\bottomrule
\end{tabular}
\label{tab:energy_prediction}
\end{table}

A clear pattern emerges in Table~\ref{tab:energy_prediction}. On ArchEGraph-P, \textbf{F2S-TransConv} performs best across all metrics, while on ArchEGraph-M, \textbf{F2S-GPS} consistently achieves the lowest errors, indicating better robustness under distribution shift.

Since all variants share the F2S backbone and differ only in attention design, the gap mainly comes from how structural dependencies are modeled: \textbf{F2S-TransConv} favors in-distribution regularity, whereas \textbf{F2S-GPS} generalizes better to irregular, manually designed geometries.

\subsection{Generalization}

We evaluate out-of-distribution generalization under two structured shifts in ArchEGraph: cross-building transfer and cross-climate transfer. Detailed subset splits are provided in Appendix~\ref{si:dataset_split}. For a fair comparison, all metrics are reported on a normalized scale, which differs from the baseline setting.

\begin{figure}[htbp]
    \centering
    \includegraphics[width=\linewidth]{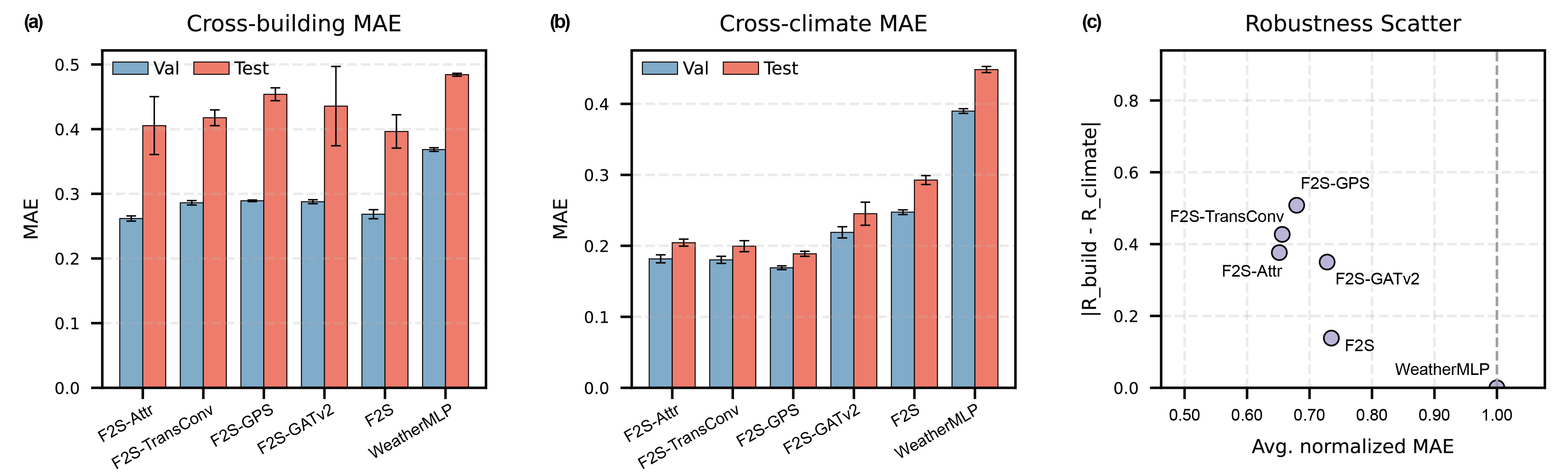}
    \caption{Cross-task generalization of energy prediction models. (a) Cross-building MAE (validation vs. test). (b) Cross-climate MAE (validation vs. test). (c) Accuracy-robustness trade-off. The x-axis shows the average normalized test MAE across both tasks; the y-axis shows cross-task inconsistency $|R_{\rm{building}}-R_{\rm{weather}}|$, where $R_{\rm{task}} = \mathrm{MAE}_{\rm{model}}^{\rm{task}} / \mathrm{MAE}_{\rm{WeatherMLP}}^{\rm{task}}$ is the test MAE of each model normalized by the WeatherMLP baseline under the same task. Lower is better in all panels.}
    \label{fig:generalization}
    \vspace{-20pt}
\end{figure}

\paragraph{Cross-building generalization}

We train on ArchEGraph-P and test on ArchEGraph-M, isolating structural distribution shift. As shown in Fig.~\ref{fig:generalization}(a), all graph-aware models outperform \textbf{WeatherMLP}, confirming that structural inductive bias remains beneficial under geometry shift. \textbf{F2S} achieves the best test MAE, while \textbf{F2S-GPS} degrades most severely despite its strong in-distribution performance.

\paragraph{Cross-climate generalization}

We train on a subset of ASHRAE climate regions and test on disjoint zones. Fig.~\ref{fig:generalization}(b) shows graph-aware models consistently outperform \textbf{WeatherMLP}, with \textbf{F2S-GPS} achieving the best test MAE. The gap against \textbf{WeatherMLP} is markedly larger than in the cross-building setting, indicating that graph structure is especially beneficial for climate generalization.

As shown in Fig.~\ref{fig:generalization}(c), the two shifts reveal a robustness trade-off: \textbf{F2S-GPS} achieves the strongest average accuracy but the highest cross-task inconsistency, while \textbf{F2S} offers the best balance. Graph topology consistently improves generalization across both shifts, with cross-building transfer remaining the harder regime. Notably, this contrasts with the in-distribution baseline results, where attention-based spatial operators (F2S-TransConv, F2S-GPS) achieved the lowest errors on large training sets; under distribution shift, physics-informed message passing with explicit opaque/transparent boundary separation (F2S, F2S-Attr) proves more robust, suggesting that domain-specific structural priors generalize better than data-driven attention when training data is limited or out-of-distribution.

\section{Limitations and future work}
\label{sec:limitation}
\vspace{-10pt}
The dataset of ArchEGraph is limited in its coverage of architectural typologies; it only incorporates office buildings, intentionally tailoring the benchmark to a single well-defined typology for controlled evaluation. The varied building geometries in the current dataset, show the robustness of the graph-based representation method and the open-source tools will allow researchers to extends the current dataset to incorporate residential and industrial buildings. Furthermore, energy simulations follow standardized construction templates and are fully simulation-based rather than measured---a deliberate choice, as building performance is inherently a forward-simulation process and consistent real-world observations at scale are unavailable at early design stages. Future work could incorporate material property variation and calibrate the data with real-world measurement to reduce simulation bias. Finally, the fixed computational budget may constrain evaluation of more expressive models; scaling to larger architectures and developing efficient graph learning methods for irregular building geometries remain open challenges. 

\section{Conclusion}
\vspace{-10pt}
We introduced ArchEGraph, a large-scale benchmark for geometry–topology–physics aligned learning for building energy modeling. By combining graph-structured building representations, aligned weather and load responses, and standardized M2G/G2E evaluation tasks, ArchEGraph supports the systematic study of how building geometry relates to energy use. The research reveals fundamental challenges in generalizing across irregular geometries and diverse climate conditions, showing the limitations of existing methods in capturing the coupling between spatial configuration and performance. We expect ArchEGraph to serve as a foundation for evaluating physically grounded learning approaches and to drive future research toward scalable, interpretable, and cross-domain models for design–simulation integration.







\bibliographystyle{plain}
\bibliography{ref.bib}

@techreport{unepb_2023,
  type = {Technical Report},
  title = {Building {{Materials}} and the {{Climate}}: {{Constructing}} a {{New Future}}},
  institution = {{United Nations Environment Programme}},
  author = {Dyson, Anna and Keena, Naomi and Lokko, Mae-ling and Reck, Barbara and Ciardullo, Christina},
  year = 2023,
  number = {978-92-807-4064-6},
  doi = {10.59117/20.500.11822/43293},
  langid = {english}
}

@article{weberHypergraphModelShows2024,
  title = {A Hypergraph Model Shows the Carbon Reduction Potential of Effective Space Use in Housing},
  author = {Weber, Ramon Elias and Mueller, Caitlin and Reinhart, Christoph},
  year = 2024,
  month = sep,
  journal = {Nature Communications},
  volume = {15},
  number = {1},
  pages = {8327},
  issn = {2041-1723},
  doi = {10.1038/s41467-024-52506-z},
}

@incollection{ashrae_climate_2021,
  title     = {Climatic Data for Building Design Standards},
  booktitle = {ASHRAE Handbook---Fundamentals},
  author    = {{ASHRAE}},
  year      = {2021},
  publisher = {American Society of Heating, Refrigerating and Air-Conditioning Engineers},
  address   = {Atlanta, GA},
  chapter   = {14}
}

@article{Kelly2015UKDALE,
  author    = {Kelly, Jack and Knottenbelt, William J.},
  title     = {The UK-DALE dataset: Domestic appliance-level electricity demand and whole-house demand from five UK homes},
  journal   = {Scientific Data},
  volume    = {2},
  pages     = {150007},
  year      = {2015},
  doi       = {10.1038/sdata.2015.7}
}

@article{Makonin2016AMPds,
  author    = {Makonin, Stephen and Ellert, Bradley and Baji{\'c}, Ivan V. and Popowich, Fred and Gill, Bram and Bartram, Linda},
  title     = {Electricity, water, and natural gas consumption of a residential house in Canada from 2012 to 2014},
  journal   = {Scientific Data},
  volume    = {3},
  pages     = {160037},
  year      = {2016},
  doi       = {10.1038/sdata.2016.37}
}

@article{douTransferLearningCrossbuilding2025,
  title = {Transfer Learning for Cross-Building Forecasting of Building Energy and Indoor Air Temperature in Model Predictive Control Applications},
  author = {Dou, Hongwen and Zhang, Kun},
  year = 2025,
  month = oct,
  journal = {Journal of Building Engineering},
  volume = {111},
  pages = {113341},
  issn = {2352-7102},
  doi = {10.1016/j.jobe.2025.113341},
  urldate = {2026-05-04},
}

@article{ribeiroTransferLearningSeasonal2018,
  title = {Transfer Learning with Seasonal and Trend Adjustment for Cross-Building Energy Forecasting},
  author = {Ribeiro, Mauro and Grolinger, Katarina and ElYamany, Hany F. and Higashino, Wilson A. and Capretz, Miriam A. M.},
  year = 2018,
  month = apr,
  journal = {Energy and Buildings},
  volume = {165},
  pages = {352--363},
  publisher = {Elsevier Science Sa},
  address = {Lausanne},
  issn = {0378-7788, 1872-6178},
  doi = {10.1016/j.enbuild.2018.01.034},
}

@article{pengClimateadaptiveEnergyForecasting2025,
  title = {Climate-Adaptive Energy Forecasting in Green Buildings via Attention-Enhanced {{Seq2Seq}} Transfer Learning},
  author = {Peng, Fang and Su, Tao and Zeng, Qing and Han, Xiaojuan},
  year = 2025,
  month = aug,
  journal = {Scientific Reports},
  volume = {15},
  number = {1},
  pages = {31829},
  issn = {2045-2322},
  doi = {10.1038/s41598-025-16953-y},
}

@article{zhangReviewMachineLearning2021,
  title = {A Review of Machine Learning in Building Load Prediction},
  author = {Zhang, Liang and Wen, Jin and Li, Yanfei and Chen, Jianli and Ye, Yunyang and Fu, Yangyang and Livingood, William},
  year = 2021,
  month = mar,
  journal = {Applied Energy},
  volume = {285},
  pages = {116452},
  issn = {0306-2619},
  doi = {10.1016/j.apenergy.2021.116452},
  urldate = {2026-01-09},
}

@article{reinhartUrbanBuildingEnergy2016a,
  title = {Urban Building Energy Modeling -- {{A}} Review of a Nascent Field},
  author = {Reinhart, Christoph F. and Cerezo Davila, Carlos},
  year = 2016,
  month = feb,
  journal = {Building and Environment},
  volume = {97},
  pages = {196--202},
  issn = {0360-1323},
  doi = {10.1016/j.buildenv.2015.12.001},
  urldate = {2026-05-04},
}

@book{clarke2001energy,
  title     = {Energy Simulation in Building Design},
  author    = {Clarke, Joseph A.},
  edition   = {2},
  year      = {2001},
  publisher = {Butterworth-Heinemann},
  address   = {Oxford, UK},
  isbn      = {978-0-7506-5082-3}
}

@misc{PecanStreet2018,
  author       = {{Pecan Street Inc.}},
  title        = {Pecan Street High-Resolution Residential Electricity Dataset},
  year         = {2018},
  note         = {40 homes, 1-second resolution},
  url          = {https://www.pecanstreet.org}
}

@misc{ASHRAEGEP2020,
  author       = {American Society of Heating, Refrigerating and Air-Conditioning Engineers},  
  title        = {ASHRAE Great Energy Predictor III Dataset},
  year         = {2020},
  howpublished = {Kaggle Competition},
  url          = {https://www.kaggle.com/competitions/ashrae-energy-prediction}
}

@article{liGeometryGraphAutomation2026,
  title = {From Geometry to Graph: {{Automation}} of Building Performance Modeling via Convex Graph Encoding},
  shorttitle = {From Geometry to Graph},
  author = {Li, Yihui and Xiao, Jun and Zhou, Hao and Lin, Borong},
  year = 2026,
  month = mar,
  journal = {Automation in Construction},
  volume = {183},
  pages = {106815},
  issn = {0926-5805},
  doi = {10.1016/j.autcon.2026.106815},
  urldate = {2026-02-04},
}

@misc{Emami2018,
  author = {Emami, Patrick and Graf, Peter},
  title = {BuildingsBench: A Large-Scale Dataset of 900K Buildings and Benchmark for Short-Term Load Forecasting},
  howpublished = {Open Energy Data Initiative (OEDI), National Renewable Energy Laboratory},
  year = {2018},
  doi = {10.25984/1986147}
}

@book{ashrae_american_society_of_heating_refrigerating_and_air-conditioning_engineers_ashrae_2017,
	address = {Atlanta, GA},
	title = {{ASHRAE} {Handbook} - {Fundamentals}},
	isbn = {978-1-939200-58-7},
	shorttitle = {{ASHRAE} {Handbook}},
	language = {English},
	publisher = {ASHRAE},
	author = {{ASHRAE American Society of Heating Refrigerating and Air-Conditioning Engineers}},
	year = {2017},
}

@article{Skeie2025,
  author = {Skeie, Kristian Stenerud and Rokseth, Lillian Sve and Lausselet, Carine and Gustavsen, Arild},
  title = {Characterizing and structuring open datasets for assessment of residential building energy use on the neighborhood and urban scale},
  journal = {Energy and Buildings},
  volume = {341},
  year = {2025},
  doi = {10.1016/j.enbuild.2025.115842}
}

@article{Pauwels2017ifcOWL,
  author = {Pauwels, Pieter and Krijnen, Thomas and Terkaj, Walter and Beetz, Jakob},
  title = {Enhancing the ifcOWL ontology with an alternative representation for geometric data},
  journal = {Automation in Construction},
  volume = {80},
  pages = {77--94},
  year = {2017},
  doi = {10.1016/j.autcon.2017.03.001}
}

@article{Janowicz2021BOT,
  author  = {Rasmussen, Mads Holten and Lefran{\c{c}}ois, Maxime and Schneider, Georg Ferdinand and Pauwels, Pieter},
  editor  = {Janowicz, Krzysztof},
  title   = {{BOT}: The Building Topology Ontology of the {W3C} Linked Building Data Group},
  journal = {Semantic Web},
  volume  = {12},
  number  = {1},
  pages   = {143--161},
  year    = {2021},
  month   = jan,
  issn    = {1570-0844},
  doi     = {10.3233/SW-200385},
}

@article{Balaji2018Brick,
  author = {Balaji, Bharathan and Bhattacharya, Arka and Fierro, Gabe and Gao, Jingkun and Gluck, Joshua and Hong, Dezhi and Johansen, Aslak and Koh, Jason and Ploennigs, Joern and Agarwal, Yuvraj and Berg{\'e}s, Mario and Culler, David and Gupta, Rajesh K. and Kj{\ae}rgaard, Mikkel Baun and Srivastava, Mani and Whitehouse, Kamin},
  title = {Brick: Metadata schema for portable smart building applications},
  journal = {Applied Energy},
  volume = {226},
  pages = {1273--1292},
  year = {2018},
  doi = {10.1016/j.apenergy.2018.02.091}
}

@misc{New2012EnergyPlus,
  author = {New, Joshua R. and Adams, Mark B. and Fricke, Brian A. and Shen, Bo and Stovall, Therese K.},
  title = {EnergyPlus},
  year = {2012},
  month =jun,
  howpublished = {Oak Ridge National Laboratory},
  url = {https://www.osti.gov/biblio/1395245}
}

@article{Mediavilla2023IFCGraphs,
  author  = {Mediavilla, Asier and Elguezabal, Peru and Lasarte, Natalia},
  title   = {Graph-Based Methodology for Multi-Scale Generation of Energy Analysis Models from {IFC}},
  journal = {Energy and Buildings},
  volume  = {282},
  pages   = {112795},
  year    = {2023},
  month   = mar,
  issn    = {0378-7788},
  doi     = {10.1016/j.enbuild.2023.112795},
}

@article{Roman2020Metamodels,
  author  = {Roman, Nadia D. and Bre, Facundo and Fachinotti, V{\'i}ctor D. and Lamberts, Roberto},
  title   = {Application and characterization of metamodels based on artificial neural networks for building performance simulation: A systematic review},
  journal = {Energy and Buildings},
  volume  = {217},
  pages   = {109972},
  year    = {2020},
  month   = jun,
  issn    = {0378-7788},
  doi     = {10.1016/j.enbuild.2020.109972},
}

@article{VonKrannichfeldt2025Hybrid,
  author = {Von Krannichfeldt, Lea and Orehounig, Kristina and Fink, Olga},
  title = {Combining physics-based and data-driven modeling for building energy systems},
  journal = {Applied Energy},
  volume = {391},
  pages = {125853},
  year = {2025},
  doi = {10.1016/j.apenergy.2025.125853}
}

@article{Scarselli2009GNN,
  author={Scarselli, Franco and Gori, Marco and Tsoi, Ah Chung and Hagenbuchner, Markus and Monfardini, Gabriele},
  journal={IEEE Transactions on Neural Networks}, 
  title={The Graph Neural Network Model}, 
  year={2009},
  volume={20},
  number={1},
  pages={61-80},
  doi={10.1109/TNN.2008.2005605}
}

@article{yuPhysicalEmbeddingBuilding2026,
  title = {Physical Embedding on Building Surface Spatial Relationships Shows Better Performance in Graph-Based Daylight Prediction},
  author = {Yu, Zhexuan and Li, Yihui and Xiao, Jun and Zhou, Hao and Lin, Borong},
  year = 2026,
  month = feb,
  journal = {Building and Environment},
  volume = {290},
  pages = {114--141},
  issn = {0360-1323},
  doi = {10.1016/j.buildenv.2025.114141},
  urldate = {2025-12-22},
}

@article{Jia2024ThermalGAT,
  author  = {Jia, Yilong and Wang, Jun and Reza Hosseini, M. and Shou, Wenchi and Wu, Peng and Mao, Chao},
  title   = {Temporal Graph Attention Network for Building Thermal Load Prediction},
  journal = {Energy and Buildings},
  volume  = {321},
  pages   = {113507},
  year    = {2024},
  month   = oct,
  issn    = {0378-7788},
  doi     = {10.1016/j.enbuild.2023.113507},
}

@article{Wu2025ResidentialBlocks,
  title = {Developing Surrogate Models for the Early-Stage Design of Residential Blocks Using Graph Neural Networks},
  author = {Wu, Zhaoji and Li, Mingkai and Liu, Wenli and Cheng, Jack C. P. and Wang, Zhe and Kwok, Helen H. L. and Huang, Cong and Hou, Fangli},
  year = 2025,
  month = jan,
  journal = {Building Simulation},
  issn = {1996-8744},
  doi = {10.1007/s12273-025-1237-7},
}

@misc{ISO16739,
  title = {Industry Foundation Classes (IFC) for data sharing in the construction and facility management industries -- Part 1: Data schema},
  organization = {International Organization for Standardization},
  author = {buildlingSMART International},
  number = {ISO 16739-1:2024},
  year = {2024},
  address = {Geneva, Switzerland},
  url = {https://www.iso.org/standard/84123.html}
}

@inproceedings{Nauata2020HouseGAN,
  author = {Nauata, Nelson and Chang, Kai-Hung and Cheng, Chin-Yi and Mori, Greg and Furukawa, Yasutaka},
  title = {House-GAN: Relational Generative Adversarial Networks for Graph-Constrained House Layout Generation},
  booktitle = {Computer Vision -- ECCV 2020},
  pages = {162--177},
  year = {2020},
  publisher = {Springer},
  doi = {10.1007/978-3-030-58452-8_10}
}

@article{Gan2022BIMGraph,
  author  = {Gan, Vincent J. L.},
  title   = {BIM-based graph data model for automatic generative design of modular buildings},
  journal = {Automation in Construction},
  volume  = {134},
  pages   = {104062},
  year    = {2022},
  doi     = {10.1016/j.autcon.2021.104062}
}

@inproceedings{Engelenburg2025MSD,
  author    = {van Engelenburg, Casper and Mostafavi, Fatemeh and Kuhn, Emanuel and Jeon, Yuntae and Franzen, Michael and Standfest, Matthias and van Gemert, Jan and Khademi, Seyran},
  title     = {{MSD}: A Benchmark Dataset for Floor Plan Generation of Building Complexes},
  booktitle = {Computer Vision -- ECCV 2024},
  year      = {2025},
  pages     = {60--75},
  publisher = {Springer},
  series    = {Lecture Notes in Computer Science},
  volume    = {15116},
  doi       = {10.1007/978-3-031-73636-0_4},
  isbn      = {978-3-031-73636-0},
}

@article{Alymani2023GraphML,
  author = {Alymani, Ali and Jabi, Wassim and Corcoran, Padraig},
  title = {Graph machine learning classification using architectural 3D topological models},
  journal = {Simulation},
  volume = {99},
  number = {11},
  pages = {1117--1131},
  year = {2023},
  doi = {10.1177/00375497221105894}
}

@article{Peters2022_3DBAG,
  author = {Peters, Ravi and Dukai, Bal{\'a}zs and Vitalis, Stelios and van Liempt, Jordi and Stoter, Jantien},
  title = {{Automated 3D Reconstruction of LoD2 and LoD1 Models for All 10 Million Buildings of the Netherlands}},
  journal = {Photogrammetric Engineering \& Remote Sensing},
  volume = {88},
  number = {3},
  pages = {165--170},
  year = {2022},
  doi = {10.14358/PERS.21-00032R2}
}

@article{Xiao2024CADBEM,
  author = {Xiao, Jun and Zhou, Hao and Yang, S. and Zhang, D. and Lin, Borong},
  title = {A CAD-BEM geometry transformation method for face-based primary geometric input based on closed contour recognition},
  journal = {Building Simulation},
  volume = {17},
  number = {2},
  pages = {335--354},
  year = {2024},
  doi = {10.1007/s12273-023-1081-6}
}

@inproceedings{Zheng2020Structured3D,
  author = {Zheng, Jia and Zhang, Junfei and Li, Jing and Tang, Rui and Gao, Shenghua and Zhou, Zihan},
  title = {Structured3D: A Large Photo-Realistic Dataset for Structured 3D Modeling},
  booktitle = {Computer Vision -- ECCV 2020},
  year = {2020},
  publisher = {Springer},
  eprint = {1908.00222},
  archivePrefix = {arXiv},
  primaryClass = {cs.CV},
  doi = {10.48550/arXiv.1908.00222}
}

@inproceedings{Cruz2021ZInD,
  author = {Cruz, Steve and Hutchcroft, Will and Li, Yuguang and Khosravan, Naji and Boyadzhiev, Ivaylo and Kang, Sing Bing},
  title = {Zillow Indoor Dataset: Annotated Floor Plans With 360deg Panoramas and 3D Room Layouts},
  booktitle = {Proceedings of the IEEE/CVF Conference on Computer Vision and Pattern Recognition (CVPR)},
  pages = {2133--2143},
  year = {2021},
  doi = {10.1109/CVPR46437.2021.00217}
}

@techreport{Deru2011DOEReference,
  author = {Deru, Michael and Field, Kristin and Studer, Daniel and Benne, Kyle and Griffith, Brent and Torcellini, Paul and Liu, Bing and Halverson, Mark and Winiarski, Dave and Rosenberg, Michael and Yazdanian, Mehry and Huang, Joe and Crawley, Drury},
  title = {{U.S. Department of Energy Commercial Reference Building Models of the National Building Stock}},
  institution = {National Renewable Energy Laboratory},
  year = {2011},
  doi = {10.2172/1009264}
}

@inproceedings{LeonSanchez2022CityGML,
  author = {Le{\'o}n-S{\'a}nchez, Camilo and Giannelli, Diego and Agugiaro, Giorgio and Stoter, Jantien},
  title = {{Creation of a CityGML-Based 3D City Model Testbed for Energy-Related Applications}},
  booktitle = {The International Archives of the Photogrammetry, Remote Sensing and Spatial Information Sciences},
  volume = {XLVIII-4/W5-2022},
  pages = {97--104},
  year = {2022},
  doi = {10.5194/isprs-archives-XLVIII-4-W5-2022-97-2022}
}

@article{Zhu2025GlobalBuildingAtlas,
  author    = {Zhu, Xiao Xiang and Chen, Sining and Zhang, Fahong and Shi, Yilei and Wang, Yuanyuan},
  title     = {GlobalBuildingAtlas: An Open Global and Complete Dataset of Building Polygons, Heights and {LoD1} {3D} Models},
  journal   = {Earth System Science Data},
  volume    = {17},
  number    = {12},
  pages     = {6647--6668},
  year      = {2025},
  month     = dec,
  publisher = {Copernicus Publications},
  issn      = {1866-3591},
  doi       = {10.5194/essd-17-6647-2025},
}

@inproceedings{Selvaraju2021BuildingNet,
  author = {Selvaraju, Pratheba and Nabail, Mohamed and Loizou, Marios and Maslioukova, Maria and Averkiou, Melinos and Andreou, Andreas and Chaudhuri, Siddhartha and Kalogerakis, Evangelos},
  title = {{BuildingNet: Learning to Label 3D Buildings}},
  booktitle = {Proceedings of the IEEE/CVF International Conference on Computer Vision},
  pages = {10397--10407},
  year = {2021},
  doi = {10.1109/ICCV48922.2021.01023}
}

@inproceedings{Wang2023Building3D,
  author = {Wang, Ruisheng and Huang, Shangfeng and Yang, Hongxin},
  title = {{Building3D: An Urban-Scale Dataset and Benchmarks for Learning Roof Structures from Point Clouds}},
  booktitle = {Proceedings of the IEEE/CVF International Conference on Computer Vision},
  pages = {20076--20086},
  year = {2023},
  doi = {10.48550/arXiv.2307.11914}
}

@article{Krapf2023PointER,
  author    = {Krapf, Sebastian and Mayer, Kevin and Fischer, Martin},
  title     = {Points for Energy Renovation ({PointER}): A Point Cloud Dataset of a Million Buildings Linked to Energy Features},
  journal   = {Scientific Data},
  volume    = {10},
  number    = {1},
  pages     = {639},
  year      = {2023},
  month     = sep,
  publisher = {Springer Nature},
  issn      = {2052-4463},
  doi       = {10.1038/s41597-023-02544-x},
}

@article{Li2021SyntheticOperation,
  author    = {Li, Han and Wang, Zhe and Hong, Tianzhen},
  title     = {A Synthetic Building Operation Dataset},
  journal   = {Scientific Data},
  volume    = {8},
  number    = {1},
  pages     = {213},
  year      = {2021},
  month     = aug,
  publisher = {Springer Nature},
  doi       = {10.1038/s41597-021-00989-6},
}

@inproceedings{Emami2023BuildingsBench,
  author    = {Emami, Patrick and Sahu, Abhijeet and Graf, Peter},
  title     = {{BuildingsBench}: A Large-Scale Dataset of 900K Buildings and Benchmark for Short-Term Load Forecasting},
  booktitle = {Advances in Neural Information Processing Systems 36 (NeurIPS 2023) -- Datasets and Benchmarks Track},
  year      = {2023},
  month     = dec,
  address   = {New Orleans, LA, USA},
  pages     = {19823--19857},
  publisher = {Neural Information Processing Systems Foundation, Inc.},
  doi       = {10.52202/075280-0871},
  arxiv     = {2307.00142},
  note      = {Datasets \& Benchmarks Track}
}

@article{Murray2017REFIT,
  author    = {Murray, David and Stankovic, Lina and Stankovic, Vladimir},
  title     = {An Electrical Load Measurements Dataset of United Kingdom Households from a Two-Year Longitudinal Study},
  journal   = {Scientific Data},
  volume    = {4},
  pages     = {160122},
  year      = {2017},
  month     = jan,
  publisher = {Springer Nature},
  doi       = {10.1038/sdata.2016.122},
}

@article{Miller2020BDG2,
  author = {Miller, Clayton and Kathirgamanathan, Anjukan and Picchetti, Bianca and Arjunan, Pandarasamy and Park, June Young and Nagy, Zoltan and Raftery, Paul and Hobson, Brodie W. and Shi, Zixiao and Meggers, Forrest},
  title = {{The Building Data Genome Project 2, Energy Meter Data from the ASHRAE Great Energy Predictor III Competition}},
  journal = {Scientific Data},
  volume = {7},
  number = {1},
  pages = {368},
  year = {2020},
  doi = {10.1038/s41597-020-00712-x}
}

@inproceedings{Zaheer2017DeepSets,
  author    = {Zaheer, Manzil and Kottur, Satwik and Ravanbakhsh, Siamak and P{\'o}czos, Barnab{\'a}s and Salakhutdinov, Ruslan R. and Smola, Alexander J.},
  title     = {Deep Sets},
  booktitle = {Advances in Neural Information Processing Systems 30 (NeurIPS 2017)},
  year      = {2017},
  month     = dec,
  address   = {Long Beach, CA, USA},
  pages     = {3394--3404},
  publisher = {Curran Associates, Inc.},
  doi       = {10.5555/3294996.3295098},
  arxiv     = {1703.06114},
}

@inproceedings{Lee2019SetTransformer,
  author    = {Lee, Juho and Lee, Yoonho and Kim, Jungtaek and Kosiorek, Adam and Choi, Seungjin and Teh, Yee Whye},
  title     = {Set Transformer: A Framework for Attention-based Permutation-Invariant Neural Networks},
  booktitle = {Proceedings of the 36th International Conference on Machine Learning (ICML)},
  year      = {2019},
  month     = jun,
  address   = {Long Beach, CA, USA},
  pages     = {3744--3753},
  publisher = {PMLR},
  series    = {Proceedings of Machine Learning Research},
  volume    = {97},
  arxiv     = {1810.00825},
  url       = {https://proceedings.mlr.press/v97/lee19d.html},
}

@inproceedings{Jaegle2021Perceiver,
  author    = {Jaegle, Andrew and Gimeno, Felix and Brock, Andy and Zisserman, Andrew and Vinyals, Oriol and Carreira, Jo{\~a}o},
  title     = {Perceiver: General Perception with Iterative Attention},
  booktitle = {Proceedings of the 38th International Conference on Machine Learning (ICML)},
  year      = {2021},
  month     = jul,
  address   = {Virtual Event},
  pages     = {4651--4664},
  publisher = {PMLR},
  series    = {Proceedings of Machine Learning Research},
  volume    = {139},
  url       = {https://proceedings.mlr.press/v139/jaegle21a.html},
}

@inproceedings{Vaswani2017Attention,
  author    = {Vaswani, Ashish and Shazeer, Noam and Parmar, Niki and Uszkoreit, Jakob and Jones, Llion and Gomez, Aidan N. and Kaiser, {\L}ukasz and Polosukhin, Illia},
  title     = {Attention Is All You Need},
  booktitle = {Advances in Neural Information Processing Systems 30 (NeurIPS 2017)},
  year      = {2017},
  month     = dec,
  address   = {Long Beach, CA, USA},
  pages     = {5998--6008},
  publisher = {Curran Associates, Inc.},
  arxiv     = {1706.03762},
  url       = {https://papers.nips.cc/paper_files/paper/2017/hash/3f5ee243547dee91fbd053c1c4a845aa-Abstract.html}
}

@inproceedings{Brody2022HowAttentive,
  author    = {Brody, Shaked and Alon, Uri and Yahav, Eran},
  title     = {How Attentive are Graph Attention Networks?},
  booktitle = {Proceedings of the Tenth International Conference on Learning Representations (ICLR)},
  year      = {2022},
  month     = apr,
  publisher = {OpenReview.net},
  url       = {https://openreview.net/forum?id=F72ximsx7C1},
  arxiv     = {2105.14491},
  doi       = {10.48550/arXiv.2105.14491},
  note      = {Virtual Event}
}

@inproceedings{Shi2021MaskedLabelPrediction,
  title     = {Masked Label Prediction: Unified Message Passing Model for Semi-Supervised Classification},
  author    = {Shi, Yunsheng and Huang, Zhengjie and Feng, Shikun and Zhong, Hui and Wang, Wenjing and Sun, Yu},
  booktitle = {Proceedings of the Thirtieth International Joint Conference on Artificial Intelligence (IJCAI-21)},
  year      = {2021},
  month     = aug,
  pages     = {1548--1554},
  publisher = {International Joint Conferences on Artificial Intelligence Organization},
  doi       = {10.24963/ijcai.2021/214},
  note      = {Main Track}
}

@inproceedings{Rampasek2022GraphGPS,
    author = {Ramp\'{a}\v{s}ek, Ladislav and Galkin, Mikhail and Dwivedi, Vijay Prakash and Luu, Anh Tuan and Wolf, Guy and Beaini, Dominique},
    title = {Recipe for a general, powerful, scalable graph transformer},
    year = {2022},
    isbn = {9781713871088},
    publisher = {Curran Associates Inc.},
    address = {Red Hook, NY, USA},
    booktitle = {Proceedings of the 36th International Conference on Neural Information Processing Systems},
    articleno = {1054},
    numpages = {15},
    location = {New Orleans, LA, USA},
    series = {NIPS '22}
}

@article{Bhandari2012Weather,
  title     = {Evaluation of weather datasets for building energy simulation},
  author    = {Bhandari, Mahabir S. and Shrestha, Som S. and New, Joshua Ryan},
  journal   = {Energy and Buildings},
  volume    = {49},
  pages     = {109--118},
  year      = {2012},
  month     = {Jun},
  publisher = {Elsevier},
  doi       = {10.1016/j.enbuild.2012.01.033}
}

@misc{xiaoFullyAutomatedDMBIMBEM2026,
  title = {A {{Fully Automated DM-BIM-BEM Pipeline Enabling Graph-Based Intelligence}}, {{Interoperability}}, and {{Performance-Driven Early Design}}},
  author = {Xiao, Jun and Wang, Qiong and Li, Yihui and Yu, Zhexuan and Zhou, Hao and Lin, Borong},
  year = 2026,
  month = jan,
  number = {arXiv:2601.16813},
  eprint = {2601.16813},
  primaryclass = {stat},
  publisher = {arXiv},
  doi = {10.48550/arXiv.2601.16813},
  urldate = {2026-02-14},
}

@misc{onebuilding_epw,
  author       = {{OneBuilding.org}},
  title        = {EnergyPlus Weather Data (EPW) Repository},
  year         = {2024},
  url          = {https://climate.onebuilding.org},
  note         = {Accessed: 2026-05-03}
}

@techreport{wilcox2008tmy3,
  author      = {Wilcox, Stephen and Marion, William},
  title       = {Users Manual for TMY3 Data Sets},
  institution = {National Renewable Energy Laboratory (NREL)},
  year        = {2008},
  number      = {NREL/TP-581-43156},
  url         = {https://www.nrel.gov/docs/fy08osti/43156.pdf}
}

@article{wangCrossplatformOptimizationSystem2025,
  title = {A Cross-Platform Optimization System for Comparative Design Exploration of Competing Concepts and Strategies},
  author = {Wang, Likai and De Luca, Francesco and Janssen, Patrick and Bui, Do Phuong Tung and Chen, Kian Wee and Yuan, Chao},
  year = 2025,
  month = dec,
  journal = {Journal of Building Engineering},
  volume = {115},
  pages = {114413},
  issn = {2352-7102},
  doi = {10.1016/j.jobe.2025.114413},
  urldate = {2026-05-04},
}

@book{edition1993ashrae,
  title     = {1993 ASHRAE Handbook: Fundamentals},
  author    = {{American Society of Heating, Refrigerating and Air-Conditioning Engineers}},
  year      = {1993},
  publisher = {ASHRAE},
  address   = {Atlanta, GA, USA},
  edition   = {SI ed.},
  isbn      = {978-0-910110-97-6}
}

@article{thom1954rational,
  title = {The Rational Relationship between Heating Degree Days and Temperature},
  author = {Thom, Herbert C. S.},
  year = {1954},
  journal = {Monthly Weather Review},
  volume = {82},
  number = {1},
  pages = {1--6},
  doi = {10.1175/1520-0493(1954)082<0001:TRRBHD>2.0.CO;2},
}

@article{quayle1980heating,
  title={Heating degree day data applied to residential heating energy consumption},
  author={Robert G. Quayle and Henry F. Diaz},
  journal={Journal of Applied Meteorology and Climatology},
  volume={19},
  number={3},
  pages={241--246},
  year={1980},
  doi={10.1175/1520-0450(1980)019<0241:HDDDAT>2.0.CO;2}
}

\newpage
\appendix

\renewcommand{\thefigure}{\Alph{section}.\arabic{figure}}
\renewcommand{\thetable}{\Alph{section}.\arabic{table}}
\renewcommand{\theequation}{\Alph{section}.\arabic{equation}}

\setcounter{figure}{0}
\setcounter{table}{0}
\setcounter{equation}{0}

\section*{Appendix}

This appendix provides additional details that support the main paper in four aspects: dataset construction and statistics, benchmark implementation and evaluation, and qualitative visualization resources. Section~\ref{si:dataset_details} expands the data pipeline, cleaning strategy, split protocol, and supplementary geometry-weather-energy analyses. Section~\ref{si:eval} reports computational settings, metric definitions, model/training configurations, and extended generalization results for both benchmark tasks. Section~\ref{si:visual} presents interactive and batch visualizations to facilitate qualitative inspection of aligned geometry-topology-physics samples. Section~\ref{si:impact} provides the boarder impacts of this research.

\section{Supplementary dataset details}
\label{si:dataset_details}

\subsection{Representative Data Case}

\begin{figure}[htbp]
    \centering
    \includegraphics[width=\linewidth]{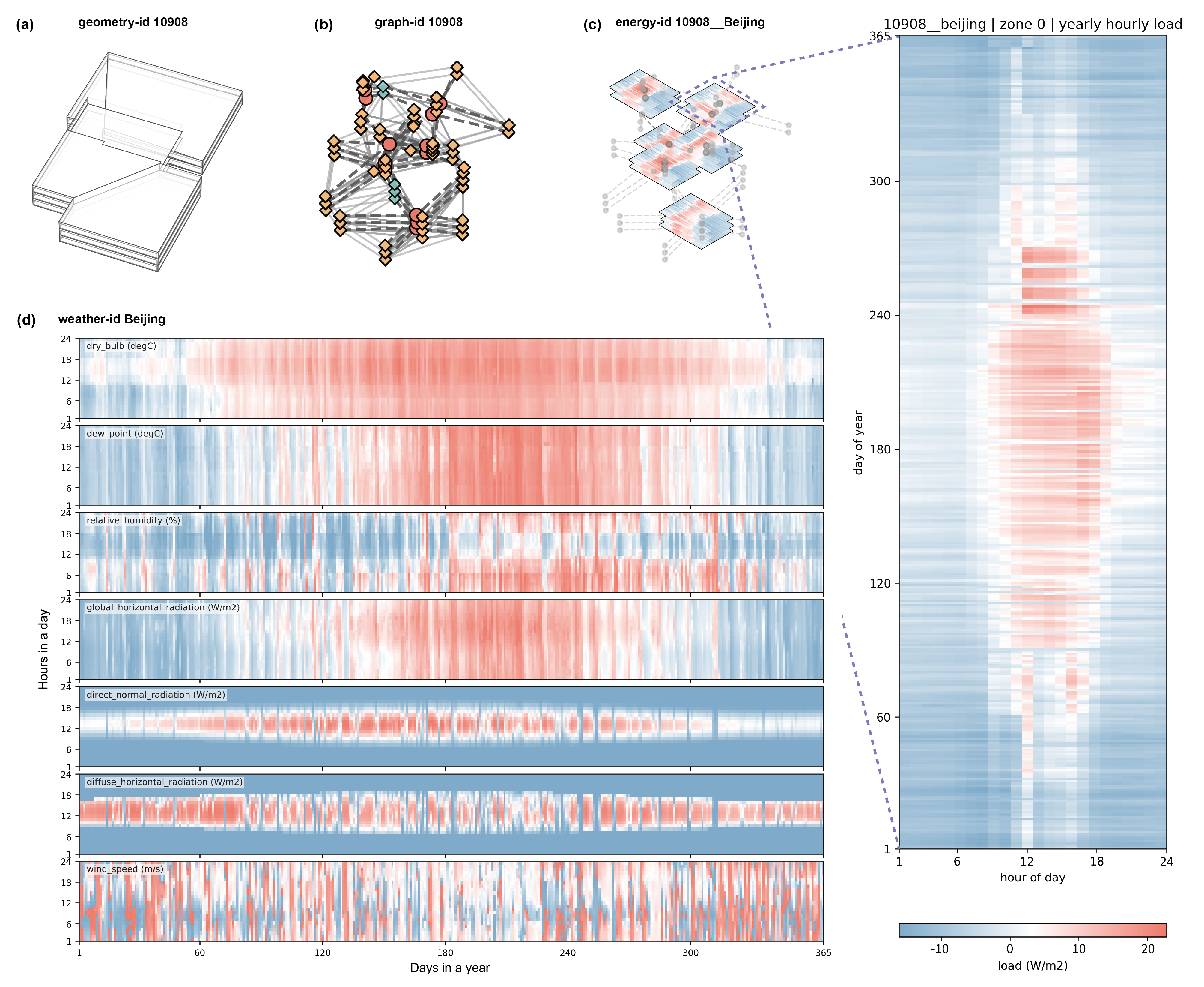}
    \caption{Example of an ArchEGraph sample showing the aligned representation of building geometry, graph topology, zone-level energy load, and local weather time series. The right panel visualizes the annual hourly load profile of a representative zone, illustrating the spatiotemporal coupling between building structure, weather conditions, and energy demand.}
    \label{fig:sample_one}
\end{figure}

Figure~\ref{fig:sample_one} shows a randomly selected data case.

\subsection{Data Preprocessing and Cleaning}
\label{si:data_pipeline}

\begin{figure}[htbp]
    \centering
    \includegraphics[width=\textwidth]{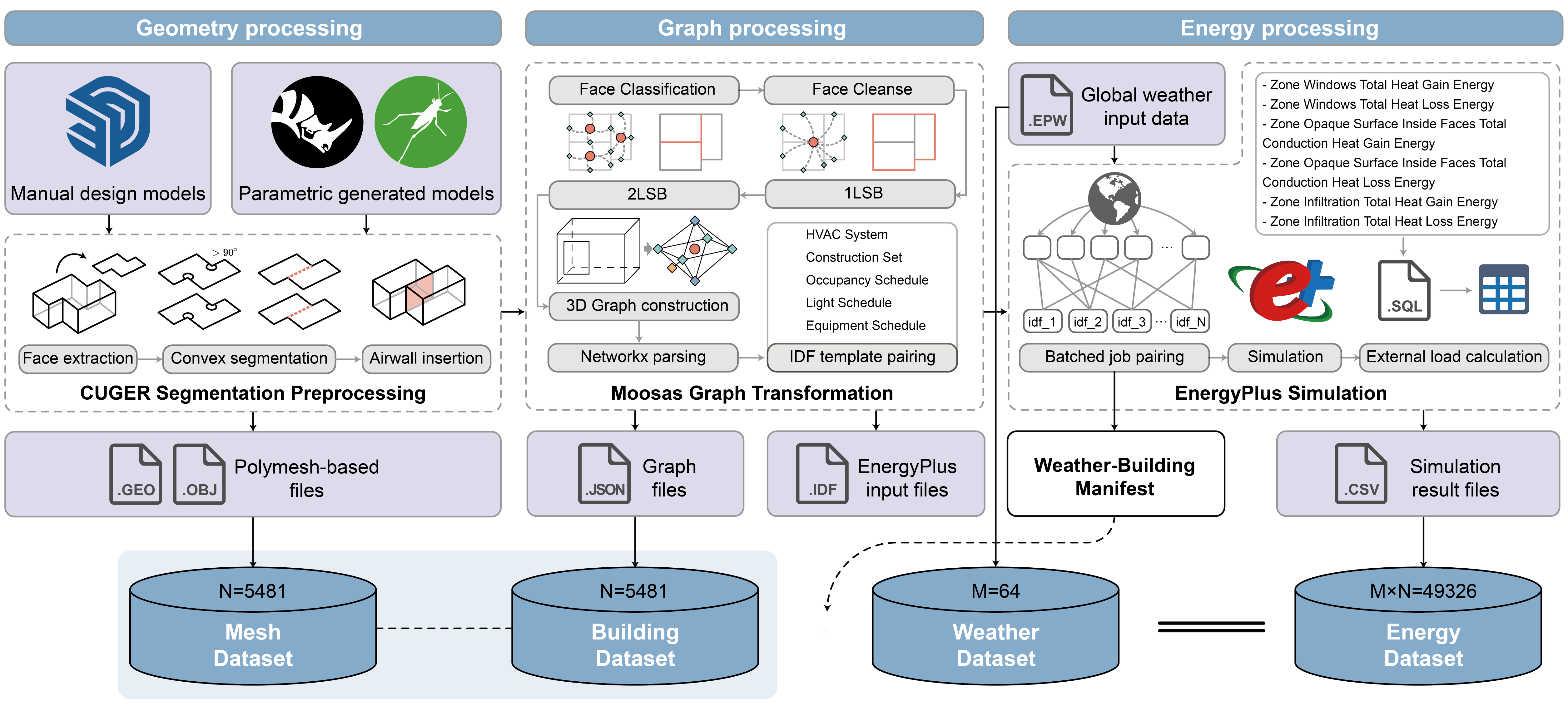}
    \caption{Dataset construction pipeline}
    \label{fig:detailed_pipeline}
\end{figure}

The pipeline consists of three sequential stages: (i) geometry normalization and convex decomposition, (ii) geometry-to-graph conversion, and (iii) physics-based simulation integration. An overview is shown in Figure~\ref{fig:detailed_pipeline}.

\subsubsection{Original Data Sources}

The geometric data in ArchEGraph are stored in OBJ format and exported from SketchUp and Rhinoceros 3D. All building instances correspond to office building typologies.

The dataset is constructed from two complementary sources:

\textbf{(1) Manually designed models.} These models are created by students in a building performance simulation and optimization course. Each design is produced under functional and spatial constraints, reflecting realistic architectural decision-making processes.

\textbf{(2) Parametric generative models.} These models are generated in Rhinoceros 3D using the EvoMass plugin~\cite{wangCrossplatformOptimizationSystem2025}, which enables controllable procedural generation of multi-story building forms. The generative process is driven by key design parameters, including floor height, column spacing, and volumetric operations (e.g., additive and subtractive transformations), allowing systematic exploration of the architectural design space.

In total, the dataset comprises 652 manually designed models and 5,052 parametric models.

\textbf{Weather data.} Climate inputs are obtained from the OneBuilding database ~\cite{onebuilding_epw}. We select 64 representative cities covering major global climate regions. For each location, the typical meteorological year (TMY) weather files are used, which are standard inputs in the building energy simulation and ensure the comparability between climates~\cite{wilcox2008tmy3}.

\subsubsection{Geometry normalization and convex decomposition}

Raw building models harvested from CAD environments (e.g., SketchUp, Rhino) typically suffer from topological defects, including duplicated vertices, inconsistent normal orientations, and non-convex boundaries. We address these through a three-stage normalization process.

\textbf{(1) Topology cleaning and vertex merging:} We first unify the geometric representation by merging vertices within a numerical tolerance $\epsilon$ to resolve leak boundaries. All faces are re-indexed to a global vertex list to eliminate redundancy, and inconsistent normals are unified via a breadth-first traversal of face adjacency to ensure a consistent inward/outward orientation.

\textbf{(2) Minimal convex partitioning:} To satisfy the requirements of both EnergyPlus simulation and graph-based spatial reasoning, all non-convex polygons (e.g., L-shaped floors or walls with openings) must be decomposed. We implement a minimal convex decomposition algorithm: complex polygons are first triangulated, followed by a local merging procedure that reconstructs the minimum number of convex sub-polygons. This step ensures that each face primitive is a simple convex polygon, facilitating stable normal calculation and centroid-based spatial indexing.

\textbf{(3) Air-wall insertion:} For models with split polygons, we introduce virtual air-wall connectors to bridge small gaps, ensuring that partially defined spatial volumes are correctly identified as watertight enclosures for simulation.

\subsubsection{Geometry-to-graph conversion}

\begin{wrapfigure}[25]{r}{0.48\textwidth}
    \centering
    \vspace{-18pt}
    \includegraphics[width=\linewidth]{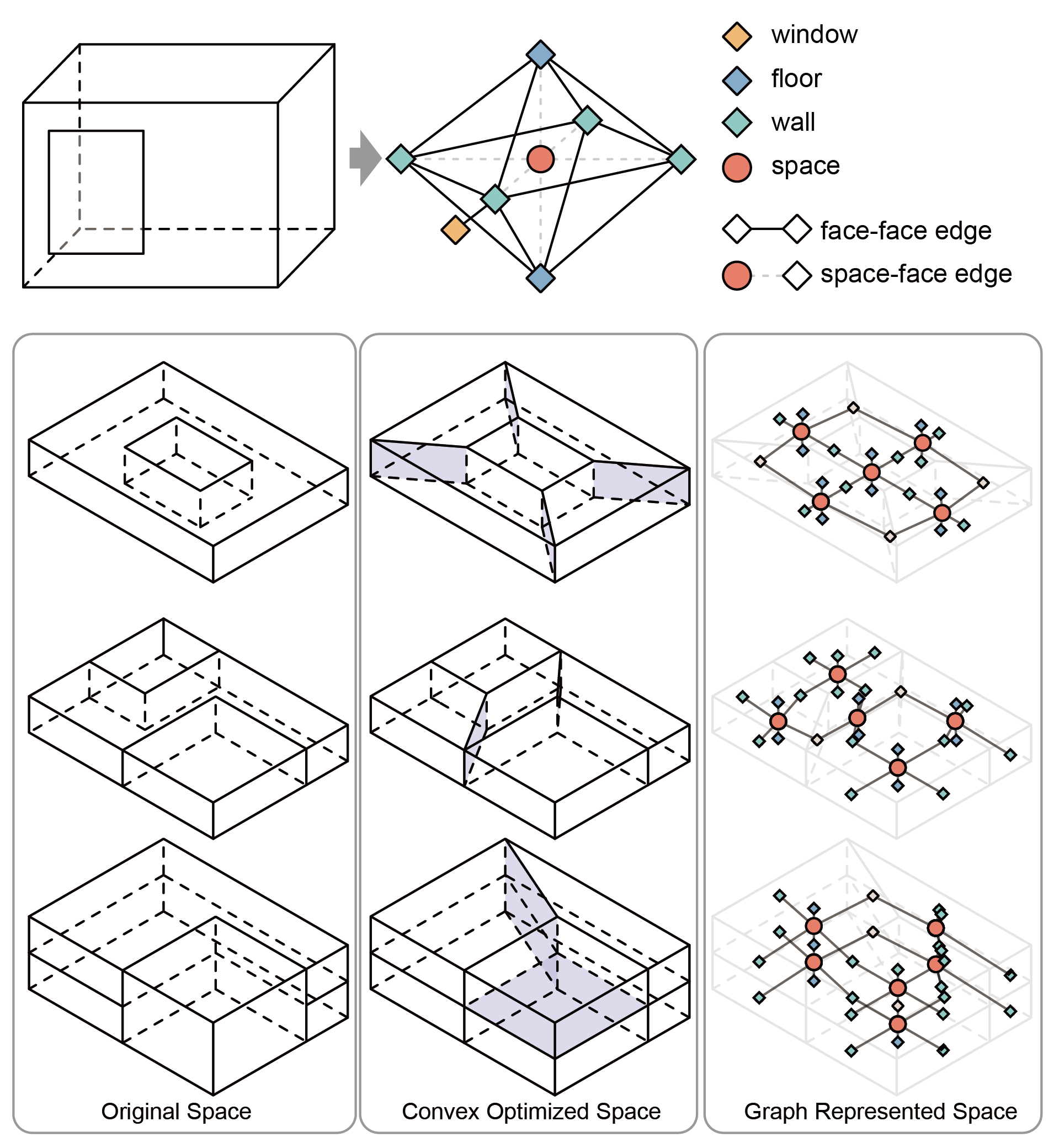}
    \caption{\textbf{Space-to-graph conversion and convex optimization pipeline.}
    \emph{(Top)} A heterogeneous graph representation: boundary surfaces are mapped to face nodes $\mathcal{V}_F$, and enclosed zones to space nodes $\mathcal{V}_S$. Edges $\mathcal{E}_{FF}$ and $\mathcal{E}_{fs}$ encode adjacency and enclosure relations.
    \emph{(Bottom)} The full pipeline from raw polygonal meshes to convex-optimized geometry and then to a heterogeneous graph representation, following the framework in~\cite{liGeometryGraphAutomation2026}.}
    \label{fig:CUGER_concept}
\end{wrapfigure}

Based on the foundational representation framework proposed in~\cite{liGeometryGraphAutomation2026}, we convert the normalized geometry $G$ into a heterogeneous graph $G = (\mathcal{V}_F, \mathcal{V}_S, \mathcal{E}_{FF}, \mathcal{E}_{FS})$. The conversion involves three critical alignment steps:

\textbf{(1) Topological adjacency detection:} Adjacency between convexified face nodes $\mathcal{V}_F$ is determined by shared boundary segments. A connectivity edge $e_{ij} \in \mathcal{E}_{FF}$ is defined when the intersection length between two faces exceeds a threshold $\tau_e$, which filters out spurious connections caused by geometric noise or numerical precision errors in CAD exports.

\textbf{(2) Volumetric enclosure reconstruction:} Following the methodology in~\cite{Xiao2024CADBEM}, we reconstruct spatial nodes $\mathcal{V}_S$ (thermal zones) by identifying closed face cycles. 

\textbf{(3) Topological refinement:} We apply a rule-based First/Second-Level Space Boundaries procedure, implemented in the open-source software MOOSAS+~\cite{xiaoFullyAutomatedDMBIMBEM2026}, to resolve ambiguities between face geometry and space semantics. The algorithm enforces a strict many-to-one mapping from faces to spatial enclosures, except when a face is explicitly classified as an interior partition shared by two spatial nodes.

The final graph integrates semantic attributes (e.g., wall, floor, roof, window) derived from CAD layer metadata and geometric heuristics. Spatial nodes aggregate their constituent faces to store high-level properties such as total volume and envelope exposure. The resulting structure, illustrated in Figure~\ref{fig:CUGER_concept}, provides a unified multi-modal representation for learning-based building performance modeling.

\subsubsection{Energy simulation integration}

The graph $G$ is deterministically serialized into EnergyPlus-compatible IDF files through a loss-preserving mapping from graph entities to thermal zones, boundary surfaces, and construction assemblies.  
Each IDF is paired with an EPW weather file selected from 64 global locations that span all ASHRAE climate regions, thereby providing broad variation in external boundary conditions.

Each building model is simulated using EnergyPlus~v24.2 with a unified set of physical and operational parameters, following the DOE Medium Office reference building standard~\cite{Deru2011DOEReference}.  
The simulation covers a full calendar year (Jan.~1--Dec.~31, 2006) at a 15-minute timestep (4 steps/hour), with solar distribution set to \texttt{FullExteriorWithReflections} and shading recalculated every 30 timesteps.  
Internal loads, envelope constructions, and HVAC settings are held constant across all building instances; only geometry and weather inputs vary across cases.  
Key parameter values are summarized in Table~\ref{tab:sim_config}.

\begin{table}[htbp]
\centering
\caption{Summary of fixed EnergyPlus simulation parameters applied uniformly to all buildings in the dataset.}
\label{tab:sim_config}
\small
\begin{tabular}{llr}
\toprule
\textbf{Category} & \textbf{Parameter} & \textbf{Value} \\
\midrule
\multirow{3}{*}{Simulation}
  & Timestep            & 4 steps/hour (15~min) \\
  & Run period          & Full year (Jan--Dec) \\
  & Solar distribution  & FullExteriorWithReflections \\
\midrule
\multirow{3}{*}{Internal loads}
  & Occupancy density   & 0.0565~people/m$^{2}$ \\
  & Lighting power density & 10.55~W/m$^{2}$ \\
  & Equipment power density & 10.33~W/m$^{2}$ \\
\midrule
\multirow{2}{*}{Infiltration}
  & Method              & Flow/Zone (constant-term only) \\
  & Rate                & $2.266\times10^{-4}$~m$^{3}$/s per m$^{2}$ exterior surface \\
\midrule
\multirow{4}{*}{Thermostat}
  & Type                & Dual setpoint (heating + cooling) \\
  & Heating setpoint (occupied) & 20--21\,\textdegree C \\
  & Cooling setpoint (occupied) & 24--25\,\textdegree C \\
  & Setback (unoccupied) & 15.6\,\textdegree C / 26.7\,\textdegree C \\
\midrule
\multirow{4}{*}{HVAC system}
  & Type                & Ideal Loads Air System (per zone) \\
  & Heating supply air  & $\leq$50\,\textdegree C, autosized capacity \\
  & Cooling supply air  & $\geq$13\,\textdegree C, autosized flow \& capacity \\
  & Economizer          & Differential dry-bulb \\
\midrule
\multirow{3}{*}{Envelope}
  & Exterior wall       & Brick + LW concrete + 50~mm insulation + gypsum board \\
  & Roof                & Membrane + 50~mm insulation + LW concrete \\
  & Window glazing      & Simple glazing: $U$=1.8~W/m$^{2}$K, SHGC=0.40, $\tau_{\mathrm{vis}}$=0.74 \\
\midrule
\multirow{2}{*}{Schedules}
  & Occupancy profile   & DOE Medium Office (peak 0.95 on weekdays) \\
  & Lighting / equipment & DOE Medium Office fractional schedules \\
\bottomrule
\end{tabular}
\end{table}

After simulation, we derive the external disturbance load, i.e., the thermal response driven by outdoor environmental forcing under fixed internal schedules (occupancy, lighting, and equipment).  
External disturbance loads represent envelope-mediated heat exchange induced by outdoor climate conditions, including solar radiation, outdoor air temperature, wind pressure, and humidity-related enthalpy effects.  
In our pipeline, these effects are computed implicitly by EnergyPlus from geometry-dependent surface exposure and material properties.  
We denote the external disturbance load at time $t$ as $Q_{\mathrm{ext}}(t)$.

Internal heat gains arise from occupants, lighting, and equipment:
\begin{equation}
Q_{\mathrm{int}}(t)=Q_{\mathrm{people}}(t)+Q_{\mathrm{lighting}}(t)+Q_{\mathrm{equipment}}(t).
\end{equation}

The total zone thermal load is then:
\begin{equation}
Q_{\mathrm{total}}(t)=Q_{\mathrm{ext}}(t)+Q_{\mathrm{int}}(t).
\end{equation}

However, building energy consumption is not identical to thermal load. HVAC systems respond to the mismatch between thermal demand and comfort targets under operational constraints. A simplified relation is:
\begin{equation}
E_{\mathrm{HVAC}}(t)\propto\left|Q_{\mathrm{total}}(t)-Q_{\mathrm{set}}(t)\right|,
\end{equation}
where $Q_{\mathrm{set}}(t)$ denotes the thermal regulation target required to maintain the prescribed comfort range.

In this dataset, we focus on $Q_{\mathrm{ext}}(t)$ as the climate-driven response signal.  
This choice is consistent with the first two stages of the pipeline, where geometry determines envelope exposure and thereby modulates external thermal forcing.  
By isolating $Q_{\mathrm{ext}}(t)$, we reduce confounding effects from HVAC control policies and internal behavioral uncertainty, yielding a cleaner setting for physics-consistent geometric representation learning.

\subsubsection{Data cleaning and implementation details}

We apply a multi-stage pipeline to ensure geometric validity, simulation reliability, and cross-modal consistency between geometry, graph structure, and simulation outputs.

\textbf{Geometric and simulation filtering.}
We first remove geometries with more than 10,000 faces, as these typically indicate corrupted models or failures in convex decomposition. We then discard cases that either fail EnergyPlus simulation or exceed a runtime of one hour, which usually corresponds to non-watertight geometry or invalid IDF conversion. Finally, we enforce consistency between simulation outputs and graph representations by computing a spatial matching ratio between detected thermal zones and graph nodes. Cases with a ratio below 80\% are excluded. After filtering, the dataset contains 580 manually designed models and 4,901 parametric models, yielding 5,481 valid graph-structured samples.

\textbf{Manifest-driven integrity and loading consistency.}
All samples are indexed via a manifest containing \texttt{sample\_id}, \texttt{building\_id}, \texttt{weather\_id}, and file paths for geometry and simulation outputs. During data loading, each entry is validated as a complete building--weather--energy triplet.

\textbf{Schema normalization and temporal alignment.}
Weather inputs stored in \texttt{npz} format are converted into a unified tabular schema. Feature aliases (e.g., different naming conventions for dry-bulb temperature) are standardized into a consistent namespace. If temporal indexing is missing, we reconstruct an hourly index to ensure deterministic sequence alignment across all samples.

\textbf{Graph-energy alignment.}
Zone-level energy signals are aligned with graph spatial nodes using a structured mapping procedure. When available, we match nodes using \texttt{valid\_energy\_spaces}. If the energy dimension exceeds graph space dimension, redundant channels are truncated; if it is smaller, zero-padding is applied. This guarantees consistent tensor shapes across all samples.

\subsection{Dataset Split}
\label{si:dataset_split}

Table~\ref{tab:bias_splits_clean} summarizes all datasets and evaluation splits used in this work. Unless otherwise specified, all standard splits follow a 70\% / 15\% / 15\% train/validation/test ratio, randomly sampled from the full dataset.

The dataset consists of two large-scale subsets, \textbf{ArchEGraph-P} and \textbf{ArchEGraph-M}, covering diverse building geometries and weather conditions. For reproducibility and lightweight experimentation, we further construct a reduced subset, \textbf{ArchEGraph-Demo}, which contains 75 buildings sampled from both P and M, each paired with 4 representative weather conditions across different climate regions.

In addition to the full datasets, we provide mesh-level splits for the M2G task (\textbf{ArchEGraph-P/M/Demo-Mesh}), which focus exclusively on geometric features without climatic variations, ensuring that specific geometries are strictly separated between sets to prevent data leakage.

To evaluate generalization under controlled conditions, we further design two bias-oriented splits:

\textbf{Building-Bias split} isolates geometric variability by fixing a single weather condition (Chicago climate). This results in 3,000 samples (one per building), partitioned into 2,100 / 450 / 450 for training, validation, and testing. Data leakage is prevented by separating building instances across splits.

\textbf{Weather-Bias split} isolates climatic variability by fixing a limited set of buildings while varying weather conditions. We select 150 buildings (75 from ArchEGraph-P and 75 from ArchEGraph-M), each paired with 20 weather scenarios, resulting in 3,000 samples. The dataset is split into 2,100 / 450 / 450, with consistent evaluation scale across both bias settings.

\begin{table}[htbp]
\centering
\caption{Statistical summary of datasets, mesh-specific splits, and bias-oriented splits.}
\label{tab:bias_splits_clean}

\resizebox{\textwidth}{!}{
\begin{tabular}{lcccccc}
\toprule
Dataset / Split 
& \#Cases 
& \#Bldg 
& \#Wea 
& Train (\#Bldg/\#Wea) 
& Val (\#Bldg/\#Wea) 
& Test (\#Bldg/\#Wea) \\
\midrule

\textbf{ArchEGraph-P} 
& 30,658 
& 4,901 
& 64 
& 21,460 (4643 / 64) 
& 4,599 (2050 / 64) 
& 4,599 (1979 / 64) \\

\textbf{ArchEGraph-P-Mesh} 
& 4,901
& 4,901
& --
& 3,431
& 735
& 735 \\

\textbf{ArchEGraph-M} 
& 18,668 
& 580 
& 64 
& 13,067 (580 / 64)
& 2,800 (572 / 54)
& 2,801 (573 / 57) \\

\textbf{ArchEGraph-M-Mesh} 
& 580
& 580
& --
& 406 
& 87 
& 87  \\

\textbf{ArchEGraph-Demo}
& 300 
& 75 
& 48 
& 210 (74 / 44) 
& 45 (35 / 26) 
& 45 (35 / 27) \\

\textbf{ArchEGraph-Demo-Mesh}
& 300
& 300
& --
& 210 
& 45 
& 45  \\

\midrule

\textbf{Building-Bias}
& 3,000 
& 3,000 
& 1 
& 2,100 (2100 / 1) 
& 450 (450 / 1) 
& 450 (450 / 1) \\

\textbf{Weather-Bias}
& 3,000 
& 150 
& 20 
& 2,100 (150 / 14) 
& 450 (150 / 3) 
& 450 (150 / 3) \\

\bottomrule
\end{tabular}
}
\end{table}

\subsection{Additional dataset statistics}
\label{si:stat}
\subsubsection{Geometry statistics}
\label{si:geo_stat}

\begin{table}[htbp]
\centering
\caption{Cross-split comparison of geometric and topological statistics in ArchEGraph.}
\label{tab:split_comparison}
\resizebox{\textwidth}{!}{%
\begin{tabular}{p{4.2cm}rrrrrrrrr}
\toprule
\textbf{Metric} & \multicolumn{3}{c}{\textbf{ArchEGraph P (4{,}901)}} & \multicolumn{3}{c}{\textbf{ArchEGraph M (580)}} & \multicolumn{3}{c}{\textbf{ArchEGraph P+M (5{,}481)}} \\
\cmidrule(lr){2-4} \cmidrule(lr){5-7} \cmidrule(lr){8-10}
 & mean & std & sum & mean & std & sum & mean & std & sum \\
\midrule

\multicolumn{10}{l}{\textbf{Graph Structure}} \\

\hspace{0.5em}Space nodes $|\mathcal{V}_S|$       
& 24.33 & 11.45 & 119{,}265 
& 24.50 & 26.47 & 14{,}209 
& 24.35 & 13.80 & 133{,}474 \\

\hspace{0.5em}Face nodes $|\mathcal{V}_F|$        
& 253.48 & 101.80 & 1{,}242{,}324 
& 351.47 & 325.41 & 203{,}853 
& 263.86 & 146.20 & 1{,}446{,}177 \\

\hspace{1.0em}\it{Floor}
& {\color{grey}\scriptsize (46.93\%)}120.54  & 52.31 & 590{,}786
& {\color{grey}\scriptsize (27.87\%)}97.32  & 204.06 & 56{,}448
& {\color{grey}\scriptsize (44.75\%)}117.96  & 77.10 & 647{,}234 \\

\hspace{1.0em}\it{Wall}
& {\color{grey}\scriptsize (22.81\%)}55.59  & 19.43 & 272{,}435
& {\color{grey}\scriptsize (30.92\%)}101.12  & 117.46 & 58{,}651
& {\color{grey}\scriptsize (22.89\%)}60.37  & 38.80 & 331{,}086 \\

\hspace{1.0em}\it{Window}
& {\color{grey}\scriptsize (21.75\%)}53.08  & 18.77 & 260{,}162
& {\color{grey}\scriptsize (39.36\%)}144.14  & 151.43 & 79{,}708
& {\color{grey}\scriptsize (23.52\%)}62.01  & 47.60 & 339{,}870 \\

\hspace{1.0em}\it{Airwall}
& {\color{grey}\scriptsize (8.77\%)}25.00  & 18.13 & 118{,}868
& {\color{grey}\scriptsize (5.91\%)}24.95  & 77.31 & 9{,}032
& {\color{grey}\scriptsize (8.84\%)}24.99  & 30.50 & 127{,}900 \\

\hspace{0.5em}Edges $|\mathcal{E}_{FF}|$
& 784.85 & 381.51 & 3{,}846{,}552
& 953.86 & 1{,}132.97 & 553{,}239
& 802.74 & 518.35 & 4{,}399{,}791 \\

\hspace{0.5em}Edges $|\mathcal{E}_{FS}|$               
& 290.31 & 138.05 & 1{,}422{,}826
& 243.01 & 378.03 & 140{,}943
& 285.31 & 179.93 & 1{,}563{,}769 \\

\midrule
\multicolumn{10}{l}{\textbf{Topology}} \\

Density $|\mathcal{E}|/|\mathcal{V}_F|$ 
& 4.26 & 0.87 & -- 
& 3.62 & 1.31 & -- 
& 4.19 & 0.95 & -- \\

Face-to-space ratio $|\mathcal{V}_F|/|\mathcal{V}_S|$  
& 10.89 & 1.53 & -- 
& 19.19 & 25.57 & -- 
& 11.87 & 4.20 & -- \\

Complexity $|\mathcal{E}|-|\mathcal{V}_F|-|\mathcal{V}_S|$               
& 815.24 & 403.54 & -- 
& 869.96 & 1{,}229.64 & -- 
& 821.05 & 520.00 & -- \\

\bottomrule
\end{tabular}%
}
\end{table}

Table~\ref{tab:split_comparison} summarizes the geometric and topological diversity of ArchEGraph. The mean cyclomatic complexity is 821.05 (std. 520.00), ranging from 22 to 21,121, indicating substantial variation from tree-like layouts to highly loopy configurations. Edge density remains stable (mean 4.19), while the face-to-space ratio averages 11.87 (std. 4.20), capturing the granularity of geometric partitioning relative to spatial zoning.

Element composition is dominated by floor (44.75\%), wall (22.89\%), and window (23.52\%) faces, and airwalls for 8.84\%. The two splits exhibit complementary characteristics: Split~P (4,901 samples) is statistically compact (e.g., face-to-space ratio 10.89 ± 1.53), while Split~M (580 samples) exhibits heavy-tailed distributions with higher structural complexity (e.g., ratio 19.19 ± 25.57) and a markedly larger window proportion (39.36\% vs.\ 21.75\%). This combination provides both stable coverage of typical layouts and exposure to rare, high-complexity structures, supporting robust generalization.

\subsubsection{Climate and energy statistics}
\label{si:energy_stat}
\begin{figure}[htbp]
    \centering
    \includegraphics[width=\textwidth]{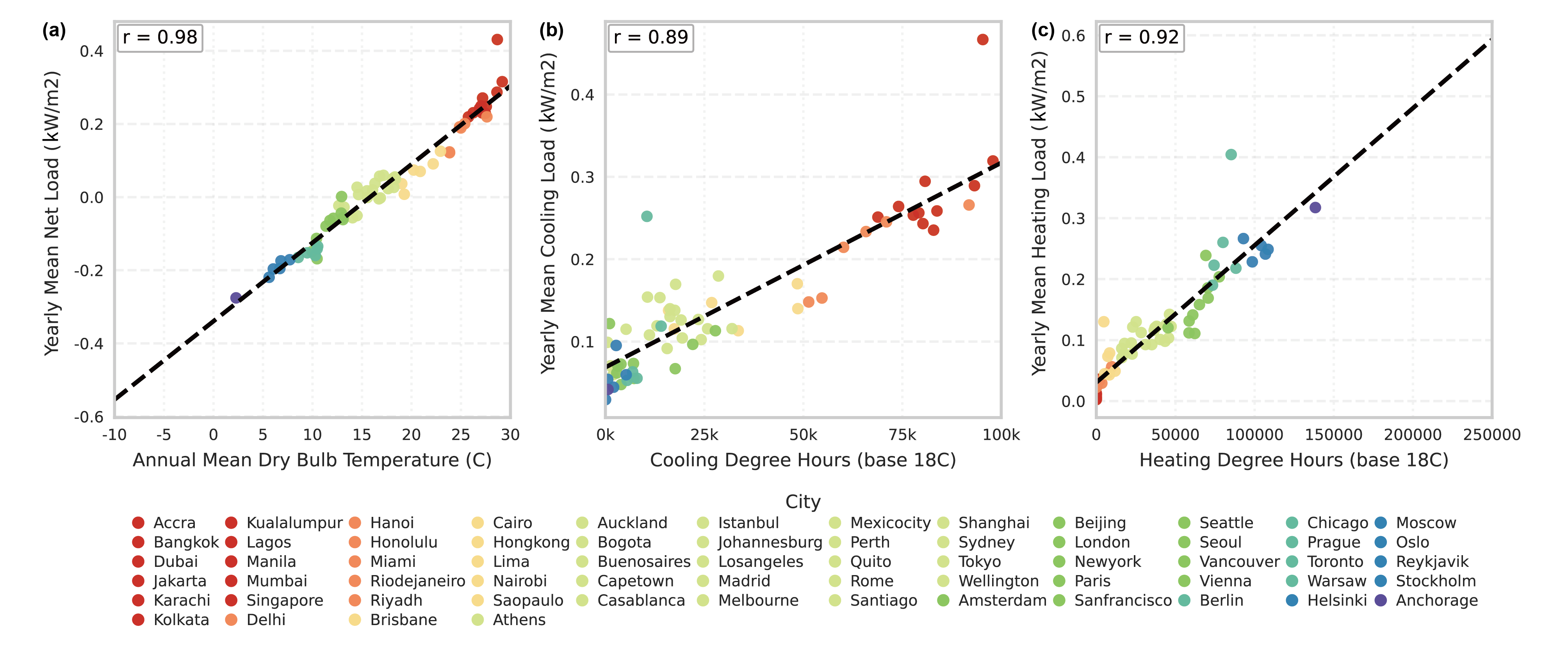}
    \caption{Climate-load regression across representative cities. From left to right: annual mean dry-bulb temperature versus annual net load, Cooling Degree Hours (base 18\,\textdegree C) versus annual cooling load, and Heating Degree Hours (base 18\,\textdegree C) versus annual heating load. Dashed lines indicate linear fits, and color denotes city identity.}
    \label{fig:weather_energy_cor}
\end{figure}

To quantify the relationship between climate drivers and energy response, Figure~\ref{fig:weather_energy_cor} shows regression results between annual building loads and key meteorological indicators, including outdoor dry-bulb temperature, Cooling Degree Days (CDD18), and Heating Degree Days (HDD18). The degree-day concept was originally introduced by Vernon L. Thom in the mid-20th century as a simple climatic index for estimating heating and cooling demand from temperature data~\cite{thom1954rational}. By accumulating deviations of daily mean temperature from a base temperature (typically 18°C), HDD and CDD provide physically interpretable measures of heating and cooling demand over time and are widely used in building energy analysis and climate zoning (e.g., ASHRAE standards~\cite{edition1993ashrae}). The results show strong linear relationships, with correlation coefficients of 0.98 for temperature vs. net load, 0.89 for CDD18 vs. cooling load, and 0.92 for HDD18 vs. heating load. Overall, energy demand is strongly governed by large-scale climatic conditions. Dry-bulb temperature captures overall load trends, while degree-day metrics better isolate heating and cooling effects by aggregating temperature deviations, reducing temporal noise and improving comparability across climates.

\begin{figure}
    \centering
    \includegraphics[width=0.75\linewidth]{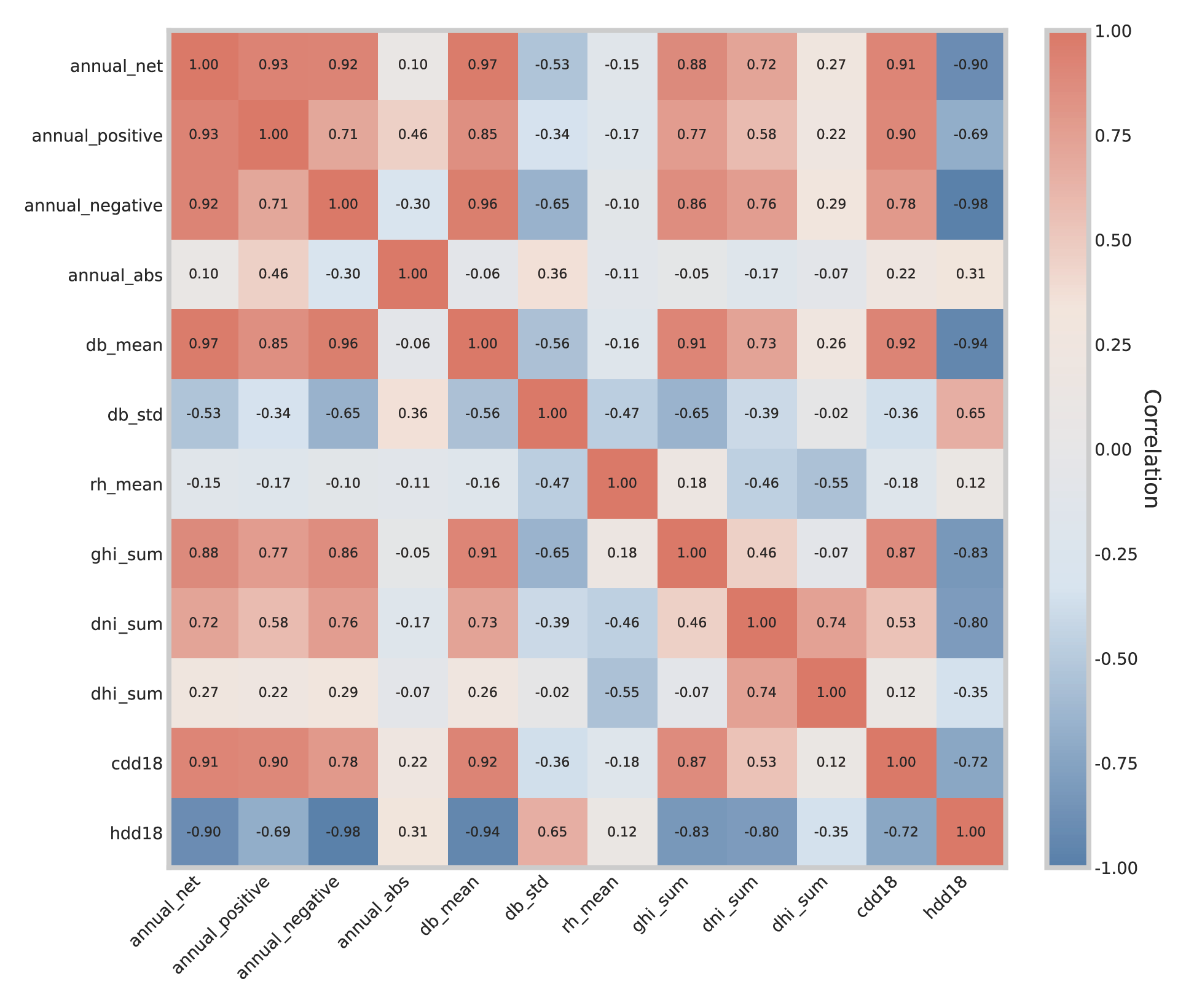}
    \caption{Correlation matrix of annual energy and weather variables.}
    \label{fig:weather_energy_heatmap}
    \vspace{-20pt}
\end{figure}

Building on these cross-city relationships, the internal dependency structure among climate and energy variables further reveals how different factors jointly shape load characteristics. As summarized in Figure~\ref{fig:weather_energy_heatmap}, annual net load is strongly consistent with its decomposed components, indicating stable load aggregation behavior. Temperature-related variables form a tightly coupled group, where mean dry-bulb temperature (db\_mean) is strongly correlated with CDD18 (0.92) and anti-correlated with HDD18 (-0.94), reflecting the separation of cooling- and heating-dominated regimes~\cite{quayle1980heating}. Solar radiation variables (e.g., ghi and dni) also show moderate to strong correlations with temperature and cooling-related metrics, suggesting their contribution to cooling demand. Overall, temperature, radiation, and degree-day metrics dominate the variation in building energy demand, while other climatic factors play a secondary role.

\section{Supplementary evaluation details}
\label{si:eval}
\subsection{Computational resources}

We conduct all experiments under a unified computational environment to ensure fair comparison across tasks. The overall evaluation protocol consists of four experimental settings: (i) M2G baseline, (ii) G2E baseline, (iii) cross-building and cross-weather generalization, and (iv) benchmark evaluation on the ArchEGraph-Demo dataset.

For each experiment, all models are trained and evaluated on the same hardware configuration. We report the average training time cost, and peak GPU memory use for each experiment. Training time is measured until convergence under early stopping, while inference time is averaged over the full test set. GPU memory usage is recorded as peak allocation during training.

The M2G and G2E experiments constitute the main computational bottleneck due to full-scale graph construction and EnergyPlus-based supervision signals, whereas the generalization and benchmark experiments are comparatively lightweight, involving fixed trained models evaluated under distribution shifts or small-scale datasets.

Overall, this setup allows us to systematically analyze computational efficiency across different stages of the proposed pipeline, ranging from geometry processing to physics-informed prediction tasks.

\begin{table}[htbp]
\centering
\caption{Computational resource usage across different experimental settings.}
\label{tab:compute_resources}
\small
\setlength{\tabcolsep}{6pt}
\resizebox{\textwidth}{!}{
\begin{tabular}{lccccc}
\toprule
Experiment & GPU Type & Case & Batch size & Training Time/h  & Peak GPU Memory/GB \\
\midrule

M2G (Geometry $\rightarrow$ Graph) 
& RTX5080(16GB) & 5k & 256 & 0.57-0.78 & 11.6-15.0 \\

G2E (Graph $\rightarrow$ Energy) 
& A100(40GB) &  30k & 8 & 5.56-20.14 & 36.8-39.0 \\

Cross-building Generalization 
& RTX5080(16GB) & 3k & 2 & 1.98-4.21  &  14.6-15.8 \\

Cross-weather Generalization 
& RTX5080(16GB) & 3k & 2 & 2.24-4.56 & 14.7-16.0 \\

ArchEGraph-Demo M2G 
& RTX5080(16GB) & 300 & 8 & 0.02-0.03 &  1.2-1.4 \\

ArchEGraph-Demo G2E 
& RTX5080(16GB) & 300 & 8 &0.18-0.22  & 12.3-15.5 \\

\bottomrule
\end{tabular}
}
\end{table}

\subsection{Model details and metric definitions}
\label{si:metric_and_model_details}

\subsubsection{M2G details}

\paragraph{Model architecture}
We design TopoTransformer to model the topological relationships between building faces and spaces. Given face features $\mathbf{x}_f\in\mathbb{R}^{N_F\times d_f}$, the model predicts three structured outputs: face-face adjacency $\mathcal{E}_{FF}$, space existence $\mathcal{V}_S$, and space-face incidence $\mathcal{E}_{FS}$.

The model first encodes face tokens using a Transformer encoder~\cite{Vaswani2017Attention}:
\begin{equation}
\mathbf{H}_F = \mathrm{Enc}_{\mathrm{Tr}}(\mathbf{x}_f),\quad \mathbf{H}_F\in\mathbb{R}^{N_F\times d_h}.
\end{equation}

Three prediction heads are then applied:

\textbf{Face-face adjacency.}
We predict pairwise adjacency using an MLP over concatenated embeddings:
\begin{equation}
\hat{z}^{FF}_{ij} = \phi_{FF}([\mathbf{h}_i \,\|\, \mathbf{h}_j]).
\end{equation}

\textbf{Space existence.}
We introduce $S_{\max}$ learnable space queries (default $64$). Each query is fused with a global pooled face representation:
\begin{equation}
\mathbf{S}=\mathrm{LN}(\mathbf{Q}+\mathbf{W}_g,\mathrm{mean}(\mathbf{H}_F)),\quad
\hat{z}^{S}_s=\phi_S(\mathbf{s}_s).
\end{equation}

\textbf{Space-face incidence.}
We compute incidence via scaled bilinear matching:
\begin{equation}
\hat{z}^{FS}_{s,i}=\frac{(\mathbf{W}_s\mathbf{s}_s)^\top(\mathbf{W}_f\mathbf{h}_i)}{\sqrt{d_h}}+b.
\end{equation}

We compare TopoTransformer against DeepSets~\cite{Zaheer2017DeepSets}, SetTransformer~\cite{Lee2019SetTransformer}, Perceiver~\cite{Jaegle2021Perceiver}, and MLP under the same M2G formulation.

\paragraph{Training objective and optimization}
The M2G training loss is:
\begin{equation}
\mathcal{L}_{\mathrm{M2G}}=\lambda_{FF}\mathcal{L}_{FF}+\lambda_{S}\mathcal{L}_{S}+\lambda_{FS}\mathcal{L}_{FS},
\end{equation}
with weights $(\lambda_{FF},\lambda_S,\lambda_{FS})=(1,1,2)$.

For $\mathcal{L}_{FF}$ and $\mathcal{L}_{FS}$, we use BCE on sampled positive/negative pairs (negative ratios 1:1 for FF and 2:1 for FS). For $\mathcal{L}_{S}$, we use a masked BCE with adaptive positive weighting to mitigate class imbalance in space-slot activation. We optimize with AdamW, gradient clipping ($\|g\|_2\le 1.0$), early stopping on validation FS-F1, and threshold 0.5 for binary metrics.

\paragraph{Default training setup}
All M2G baselines are trained under a unified setting for fair comparison. The default setup in the released code uses: Transformer hidden dimension 256, batch size 8, 30 epochs, learning rate $3\times10^{-4}$, weight decay $10^{-4}$, deterministic training, and explicit split CSV files for train/validation/test partitioning.

\paragraph{Metric definitions}
For each topology component $c\in\{\mathcal{E}_{FF},\mathcal{V}_{S},\mathcal{E}_{FS}\}$,
we binarize predicted probabilities by threshold $\tau$:
\begin{equation}
    \hat y_i^{(c)}=\mathbb{I}\!\left(\hat p_i^{(c)}\ge \tau\right).
\end{equation}

Let $\mathrm{TP}_c,\mathrm{FP}_c,\mathrm{TN}_c,\mathrm{FN}_c$ denote true positives, false positives, true negatives, and false negatives, respectively.
We report Accuracy and F1 \emph{separately} for each component:

\begin{equation}
    \mathrm{Acc}_c= \frac{\mathrm{TP}_c+\mathrm{TN}_c} {\mathrm{TP}_c+\mathrm{TN}_c+\mathrm{FP}_c+\mathrm{FN}_c},
\end{equation}

\begin{equation}
    \mathrm{F1}_c=\frac{2\,\mathrm{TP}_c}{2\,\mathrm{TP}_c+\mathrm{FP}_c+\mathrm{FN}_c}.
\end{equation}

\subsubsection{G2E details}

\paragraph{Model architecture}

The default G2E model (F2S-Attr) is a heterogeneous graph-to-sequence framework that maps building topology and weather conditions to hourly energy loads through spatial message passing and temporal decoding. The architecture operates on a face–space graph augmented with external environmental signals, and consists of three main components: (i) masked heterogeneous message passing, (ii) spatial encoding with attribute-aware edge modeling, and (iii) temporal decoding with a GRU-based sequence model.

Face and space attributes are first embedded via MLP encoders to obtain initial node representations $\mathbf{h}_f$ and $\mathbf{h}_s$. Weather variables are encoded with cyclical temporal features and contextual augmentation (see below) to produce $\mathbf{w}_t$. In F2S-Attr, weather information is injected only into exterior face nodes via a structural mask. Formally, let $\mathcal{V}_F^{\text{out}} \subset \mathcal{V}_F$ denote exterior faces; then the masked weather embedding is defined as
\begin{equation}
\mathbf{h}_f^{(0)} =
\begin{cases}
\phi_f(\mathbf{x}_f) + \phi_w(\mathbf{w}_t), & f \in \mathcal{V}_F^{\text{out}},\\
\phi_f(\mathbf{x}_f), & f \notin \mathcal{V}_F^{\text{out}},
\end{cases}
\end{equation}
where $\phi_f(\cdot)$ and $\phi_w(\cdot)$ are MLP encoders for geometric attributes and weather signals, respectively. This design ensures that weather conditions are only injected at exterior boundaries, while interior faces receive information purely through geometric propagation.

Message passing is performed on face-to-space edges with attribute-aware and physics-inspired separation between opaque and transparent interfaces. For each edge $e_{fs}$, the message is computed as
\begin{equation}
\mathbf{m}_{f \rightarrow s} =
\psi \big( [ \mathbf{h}_f \,\|\, \mathbf{e}_{fs} \,\|\, \mathbf{h}_s ] \big),
\end{equation}
where $\mathbf{e}_{fs}$ encodes geometric adjacency and material properties. To explicitly model heat-transfer heterogeneity, edges are categorized into transparent and opaque types via a binary indicator $\tau_{fs} \in \{0,1\}$. We apply a gated modulation:
\begin{equation}
\tilde{\mathbf{m}}_{f \rightarrow s} =
\alpha_{fs} \cdot \mathbf{m}_{f \rightarrow s}, \quad
\alpha_{fs} = \sigma\big( \mathbf{a}^\top \mathbf{e}_{fs} + b \big),
\end{equation}
where the attention gate $\alpha_{fs}$ is further conditioned on $\tau_{fs}$ to distinguish radiative versus conductive transfer behavior. Aggregated messages update space embeddings via residual stacking:
\begin{equation}
\mathbf{h}_s^{(l+1)} = \mathbf{h}_s^{(l)} + \sum_{f \in \mathcal{N}(s)} \tilde{\mathbf{m}}_{f \rightarrow s}.
\end{equation}

After spatial encoding, each space representation is concatenated with time-dependent weather embeddings and passed into a GRU decoder for sequential prediction:
\begin{equation}
\hat{\mathbf{y}}_t = \mathrm{GRU}\big(\mathbf{h}_s \,\|\, \mathbf{w}_t^{\text{enc}}\big).
\end{equation}

To improve temporal expressivity, the hourly index $t \in \{1, \dots, 8760\}$ is embedded using cyclical positional encoding:
\begin{equation}
\mathbf{x}_{t}^{\mathrm{time}} =
\begin{bmatrix}
\sin\left(\frac{2\pi (t-1)}{24}\right), &
\cos\left(\frac{2\pi (t-1)}{24}\right), &
\sin\left(\frac{2\pi (t-1)}{168}\right), &
\cos\left(\frac{2\pi (t-1)}{168}\right)
\end{bmatrix}^{\mathrm{T}}.
\end{equation}

Weather inputs are further augmented with short-term and seasonal context:
\begin{equation}
\mathbf{w}'_t = \left[ w_t, \{w_{t \pm h}\}_{h=1}^{3}, \{w_{t \pm 24d}\}_{d=1}^{3} \right],
\end{equation}
capturing both local fluctuations and periodic climate trends.

Finally, to stabilize training, targets are transformed into a signed-log space and normalized:
\begin{equation}
u = \mathrm{sign}(y)\cdot \ln(1 + \alpha |y|), \quad
\tilde{u} = \frac{u - \mu_u}{\sigma_u + \epsilon}.
\end{equation}

To evaluate the robustness of our architecture, we benchmark several operator variants by replacing the spatial message-passing module with GATv2 \cite{Brody2022HowAttentive}, TransConv \cite{Shi2021MaskedLabelPrediction}, and GraphGPS \cite{Rampasek2022GraphGPS}, while keeping all masking, encoding, and decoding components intact. Furthermore, we conduct an ablation study using F2S, which omits the masked encoding mechanism, and a baseline WeatherMLP model that predicts loads using only meteorological conditions.

\paragraph{Training objective and optimization}
G2E is trained with MSE loss in normalized target space, AdamW optimization, gradient clipping ($\|g\|_2\le 1.0$), and early stopping on validation RMSE.

\paragraph{Default training setup}
All G2E baselines are trained under a unified setting for fair comparison. The default setup uses batch size 8, 100 epochs, learning rate $10^{-4}$, and deterministic seeds; evaluation reports both normalized-scale metrics and inverse-transformed original-scale metrics.

\paragraph{Metric definitions}
For G2E, let $y_{s,t}$ and $\hat y_{s,t}$ denote the ground-truth and prediction
for space node $v_S \in \mathcal{V}_S$ at time step $t=1,\dots,T$.
We report node-level MAE, RMSE, and MSE:

\begin{equation}
\mathrm{MAE}=\frac{1}{|\mathcal{V}_S|}\sum_{v_S\in\mathcal{V}_S}\frac{1}{T}\sum_{t=1}^{T}\left|y_{s,t}-\hat y_{s,t}\right|,
\end{equation}

\begin{equation}
\mathrm{MSE}=\frac{1}{|\mathcal{V}_S|}\sum_{v_S\in\mathcal{V}_S}
\frac{1}{T}\sum_{t=1}^{T}(y_{s,t}-\hat y_{s,t})^2.
\end{equation}

\begin{equation}
\mathrm{RMSE}=\frac{1}{|\mathcal{V}_S|}\sum_{v_S\in\mathcal{V}_S}
\sqrt{\frac{1}{T}\sum_{t=1}^{T}(y_{s,t}-\hat y_{s,t})^2},
\end{equation}

\subsection{Additional generalization results}

We report additional generalization results to further support the robustness analysis presented in the main paper. These experiments extend the primary evaluation by testing model behavior under more challenging distribution shifts, including cross-building transfer from parametric to manually designed geometries (Table~\ref{tab:cross_building}) and cross-climate transfer across ASHRAE regions (Table~\ref{tab:cross_climate}). While not included in the main results, these settings provide a more comprehensive view of model stability under changes in spatial structure and environmental conditions.

\begin{table}[htbp]
\centering
\scriptsize
\setlength{\tabcolsep}{2pt}
\caption{Cross-building generalization from parametric (P) to manually designed (M) buildings.}
\begin{tabular}{lcccccc}
\toprule
& \multicolumn{3}{c}{Validation (P)}
& \multicolumn{3}{c}{Test (M)} \\
\cmidrule(lr){2-4} \cmidrule(lr){5-7}
Method & MAE & MSE & RMSE
& MAE & MSE & RMSE \\
\midrule
WeatherMLP        
& $ 0.3684 \,\textcolor{gray}{\scriptscriptstyle \pm 0.0029} $ 
& $ 0.4170 \,\textcolor{gray}{\scriptscriptstyle \pm 0.0013} $ 
& $ 0.6457 \,\textcolor{gray}{\scriptscriptstyle \pm 0.0010} $
& $ 0.4811 \,\textcolor{gray}{\scriptscriptstyle \pm 0.0023} $ 
& $ 0.6113 \,\textcolor{gray}{\scriptscriptstyle \pm 0.0024} $ 
& $ 0.7818 \,\textcolor{gray}{\scriptscriptstyle \pm 0.0015} $ \\

F2S
& $ 0.2684 \,\textcolor{gray}{\scriptscriptstyle \pm 0.0070} $ 
& $ 0.4292 \,\textcolor{gray}{\scriptscriptstyle \pm 0.0091} $ 
& $ 0.6551 \,\textcolor{gray}{\scriptscriptstyle \pm 0.0070} $ 
& $ \mathbf{0.3964} \,\textcolor{gray}{\scriptscriptstyle \pm 0.0258} $ 
& $ \mathbf{0.4577} \,\textcolor{gray}{\scriptscriptstyle \pm 0.0448} $ 
& $ \mathbf{0.6760} \,\textcolor{gray}{\scriptscriptstyle \pm 0.0328} $ \\

F2S-Attr           
& $ \mathbf{0.2618} \,\textcolor{gray}{\scriptscriptstyle \pm 0.0040} $ 
& $ \mathbf{0.4145} \,\textcolor{gray}{\scriptscriptstyle \pm 0.0094} $ 
& $ \mathbf{0.6438} \,\textcolor{gray}{\scriptscriptstyle \pm 0.0073} $ 
& $ 0.4055 \,\textcolor{gray}{\scriptscriptstyle \pm 0.0448} $ 
& $ 0.5361 \,\textcolor{gray}{\scriptscriptstyle \pm 0.1109} $ 
& $ 0.7296 \,\textcolor{gray}{\scriptscriptstyle \pm 0.0751} $ \\

F2S-GATv2         
& $ 0.2877 \,\textcolor{gray}{\scriptscriptstyle \pm 0.0031} $ 
& $ 0.4369 \,\textcolor{gray}{\scriptscriptstyle \pm 0.0055} $ 
& $ 0.6610 \,\textcolor{gray}{\scriptscriptstyle \pm 0.0042} $ 
& $ 0.4356 \,\textcolor{gray}{\scriptscriptstyle \pm 0.0613} $ 
& $ 0.5392 \,\textcolor{gray}{\scriptscriptstyle \pm 0.1231} $ 
& $ 0.7313 \,\textcolor{gray}{\scriptscriptstyle \pm 0.0819} $ \\

F2S-TransConv   
& $ 0.2861 \,\textcolor{gray}{\scriptscriptstyle \pm 0.0034} $ 
& $ 0.4400 \,\textcolor{gray}{\scriptscriptstyle \pm 0.0119} $ 
& $ 0.6633 \,\textcolor{gray}{\scriptscriptstyle \pm 0.0090} $ 
& $ 0.4176 \,\textcolor{gray}{\scriptscriptstyle \pm 0.0122} $ 
& $ 0.5301 \,\textcolor{gray}{\scriptscriptstyle \pm 0.0624} $ 
& $ 0.7273 \,\textcolor{gray}{\scriptscriptstyle \pm 0.0422} $ \\

F2S-GPS          
& $ 0.2892 \,\textcolor{gray}{\scriptscriptstyle \pm 0.0012} $ 
& $ 0.4450 \,\textcolor{gray}{\scriptscriptstyle \pm 0.0067} $ 
& $ 0.6671 \,\textcolor{gray}{\scriptscriptstyle \pm 0.0051} $ 
& $ 0.4539 \,\textcolor{gray}{\scriptscriptstyle \pm 0.0100} $ 
& $ 0.5553 \,\textcolor{gray}{\scriptscriptstyle \pm 0.0208} $ 
& $ 0.7451 \,\textcolor{gray}{\scriptscriptstyle \pm 0.0139} $ \\

\bottomrule
\end{tabular}
\label{tab:cross_building}
\end{table}

\begin{table}[htbp]
\centering
\scriptsize
\setlength{\tabcolsep}{2pt}
\caption{Cross-climate generalization across ASHRAE climate regions.}
\begin{tabular}{lcccccc}
\toprule
& \multicolumn{3}{c}{Validation}
& \multicolumn{3}{c}{Test} \\
\cmidrule(lr){2-4} \cmidrule(lr){5-7}
Method & MAE & MSE & RMSE
& MAE & MSE & RMSE \\
\midrule
WeatherMLP         
& $ 0.3899 \,\textcolor{gray}{\scriptscriptstyle \pm 0.0033} $ 
& $ 0.6515 \,\textcolor{gray}{\scriptscriptstyle \pm 0.0050} $ 
& $ 0.8071 \,\textcolor{gray}{\scriptscriptstyle \pm 0.0031} $ 
& $ 0.4484 \,\textcolor{gray}{\scriptscriptstyle \pm 0.0043} $ 
& $ 0.8754 \,\textcolor{gray}{\scriptscriptstyle \pm 0.0017} $ 
& $ 0.9356 \,\textcolor{gray}{\scriptscriptstyle \pm 0.0009} $ \\

F2S           
& $ 0.2474 \,\textcolor{gray}{\scriptscriptstyle \pm 0.0033} $ 
& $ 0.3195 \,\textcolor{gray}{\scriptscriptstyle \pm 0.0065} $ 
& $ 0.5652 \,\textcolor{gray}{\scriptscriptstyle \pm 0.0058} $ 
& $ 0.2927 \,\textcolor{gray}{\scriptscriptstyle \pm 0.0063} $ 
& $ 0.4625 \,\textcolor{gray}{\scriptscriptstyle \pm 0.0124} $ 
& $ 0.6800 \,\textcolor{gray}{\scriptscriptstyle \pm 0.0091} $ \\
F2S-Attr      
& $ 0.1817 \,\textcolor{gray}{\scriptscriptstyle \pm 0.0056} $ 
& $ 0.1511 \,\textcolor{gray}{\scriptscriptstyle \pm 0.0064} $ 
& $ 0.3886 \,\textcolor{gray}{\scriptscriptstyle \pm 0.0082} $ 
& $ 0.2045 \,\textcolor{gray}{\scriptscriptstyle \pm 0.0049} $ 
& $ 0.2027 \,\textcolor{gray}{\scriptscriptstyle \pm 0.0097} $ 
& $ 0.4502 \,\textcolor{gray}{\scriptscriptstyle \pm 0.0107} $ \\

F2S-GATv2          
& $ 0.2190 \,\textcolor{gray}{\scriptscriptstyle \pm 0.0079} $ 
& $ 0.2011 \,\textcolor{gray}{\scriptscriptstyle \pm 0.0208} $ 
& $ 0.4481 \,\textcolor{gray}{\scriptscriptstyle \pm 0.0232} $ 
& $ 0.2453 \,\textcolor{gray}{\scriptscriptstyle \pm 0.0163} $ 
& $ 0.2698 \,\textcolor{gray}{\scriptscriptstyle \pm 0.0411} $ 
& $ 0.5184 \,\textcolor{gray}{\scriptscriptstyle \pm 0.0400} $ \\
F2S-TransConv    
& $ 0.1803 \,\textcolor{gray}{\scriptscriptstyle \pm 0.0050} $ 
& $ 0.1412 \,\textcolor{gray}{\scriptscriptstyle \pm 0.0058} $ 
& $ 0.3758 \,\textcolor{gray}{\scriptscriptstyle \pm 0.0077} $ 
& $ 0.1995 \,\textcolor{gray}{\scriptscriptstyle \pm 0.0078} $ 
& $ 0.1906 \,\textcolor{gray}{\scriptscriptstyle \pm 0.0116} $ 
& $ 0.4364 \,\textcolor{gray}{\scriptscriptstyle \pm 0.0132} $ \\
F2S-GPS           
& $ \mathbf{0.1691} \,\textcolor{gray}{\scriptscriptstyle \pm 0.0027} $ 
& $ \mathbf{0.1347} \,\textcolor{gray}{\scriptscriptstyle \pm 0.0035} $ 
& $ \mathbf{0.3670} \,\textcolor{gray}{\scriptscriptstyle \pm 0.0047} $ 
& $ \mathbf{0.1888} \,\textcolor{gray}{\scriptscriptstyle \pm 0.0035} $ 
& $ \mathbf{0.1784} \,\textcolor{gray}{\scriptscriptstyle \pm 0.0032} $ 
& $ \mathbf{0.4223} \,\textcolor{gray}{\scriptscriptstyle \pm 0.0038} $ \\

\bottomrule
\end{tabular}
\label{tab:cross_climate}
\end{table}

\subsection{Demo dataset benchmark on two tasks}

\begin{wraptable}[20]{r}{0.63\textwidth}
\centering
\small
\setlength{\tabcolsep}{2pt}
\vspace{-12pt}
\caption{Benchmark results on the demo subset for M2G/G2E tasks.}
\label{tab:demo_both}

\begin{tabular*}{\linewidth}{@{\extracolsep{\fill}}lcccccc}
\toprule

\multicolumn{7}{c}{\textbf{M2G Task}} \\
\midrule
& \multicolumn{2}{c}{$\mathcal{E}_{FF}$} 
& \multicolumn{2}{c}{$\mathcal{V}_S$} 
& \multicolumn{2}{c}{$\mathcal{E}_{FS}$} \\
\cmidrule(lr){2-3} \cmidrule(lr){4-5} \cmidrule(lr){6-7}
Method & F1 & Acc & F1 & Acc & F1 & Acc \\
\midrule
MLP & 0.9165 & 0.9121 & 0.8605 & 0.9160 & 0.2493 & 0.8577 \\
DeepSets & 0.9546 & 0.9528 & 0.8680 & 0.9167 & 0.2947 & 0.8202 \\
Perceiver & 0.9244 & 0.9189 & 0.8575 & 0.9094 & 0.2413 & \textbf{0.8801} \\
SetTransformer & \textbf{0.9570} & \textbf{0.9553} & 0.8685 & 0.9160 & \textbf{0.3212} & 0.8460 \\
TopoTransformer & 0.9541 & 0.9522 & \textbf{0.8686} & \textbf{0.9194} & 0.3144 & 0.8538 \\
\end{tabular*}

\begin{tabular*}{\linewidth}{@{\extracolsep{\fill}}lcccccc}
\toprule
\multicolumn{7}{c}{\textbf{G2E Task}} \\
\midrule
& \multicolumn{3}{c}{Normalized} & \multicolumn{3}{c}{Original scale} \\
\cmidrule(lr){2-4} \cmidrule(lr){5-7}
Model & MAE & MSE & RMSE & MAE & MSE & RMSE \\
\midrule
WeatherMLP      & 0.5108 & 0.5713 & 0.7558 & 0.0067 & 0.000289 & 0.0170 \\
F2S             & 0.4169 & 0.4195 & 0.6477 & 0.0056 & 0.000242 & 0.0155 \\
F2S-Attr         & 0.3823 & 0.3657 & 0.6047 & 0.0052 & 0.000215 & 0.0147 \\
F2S-GATv2           & 0.4217 & 0.4182 & 0.6467 & 0.0056 & 0.000233 & 0.0153 \\
F2S-TransConv & 0.3663 & 0.3261 & 0.5710 & 0.0050 & 0.000186 & 0.0136 \\
F2S-GPS        & \textbf{0.3458} & \textbf{0.3023} & \textbf{0.5498} & \textbf{0.0046} & \textbf{0.000151} & \textbf{0.0123} \\
\bottomrule
\end{tabular*}

\end{wraptable}

To provide a compact benchmark on demo subset, we evaluate representative models on both M2G and G2E tasks in Table~\ref{tab:demo_both}. For M2G, results are reported on face–face adjacency ($\mathcal{E}_{FF}$), space-node classification ($\mathcal{V}_S$), and face–space relations ($\mathcal{E}_{FS}$), measured by F1-score and accuracy. For G2E, we compare neural and graph-based models under both normalized and original scales, enabling a consistent assessment across different magnitudes and learning settings. Overall, graph/transformer-based models consistently outperform MLP and set-based baselines across all subtasks, with stronger gains observed on more structurally complex relations such as face–space interactions.

\section{Additional visualizations}
\label{si:visual}
\subsection{Online visualization platform for demo dataset}

\begin{figure}[htbp]
    \centering
    \fbox{
        \includegraphics[width=0.99\linewidth]{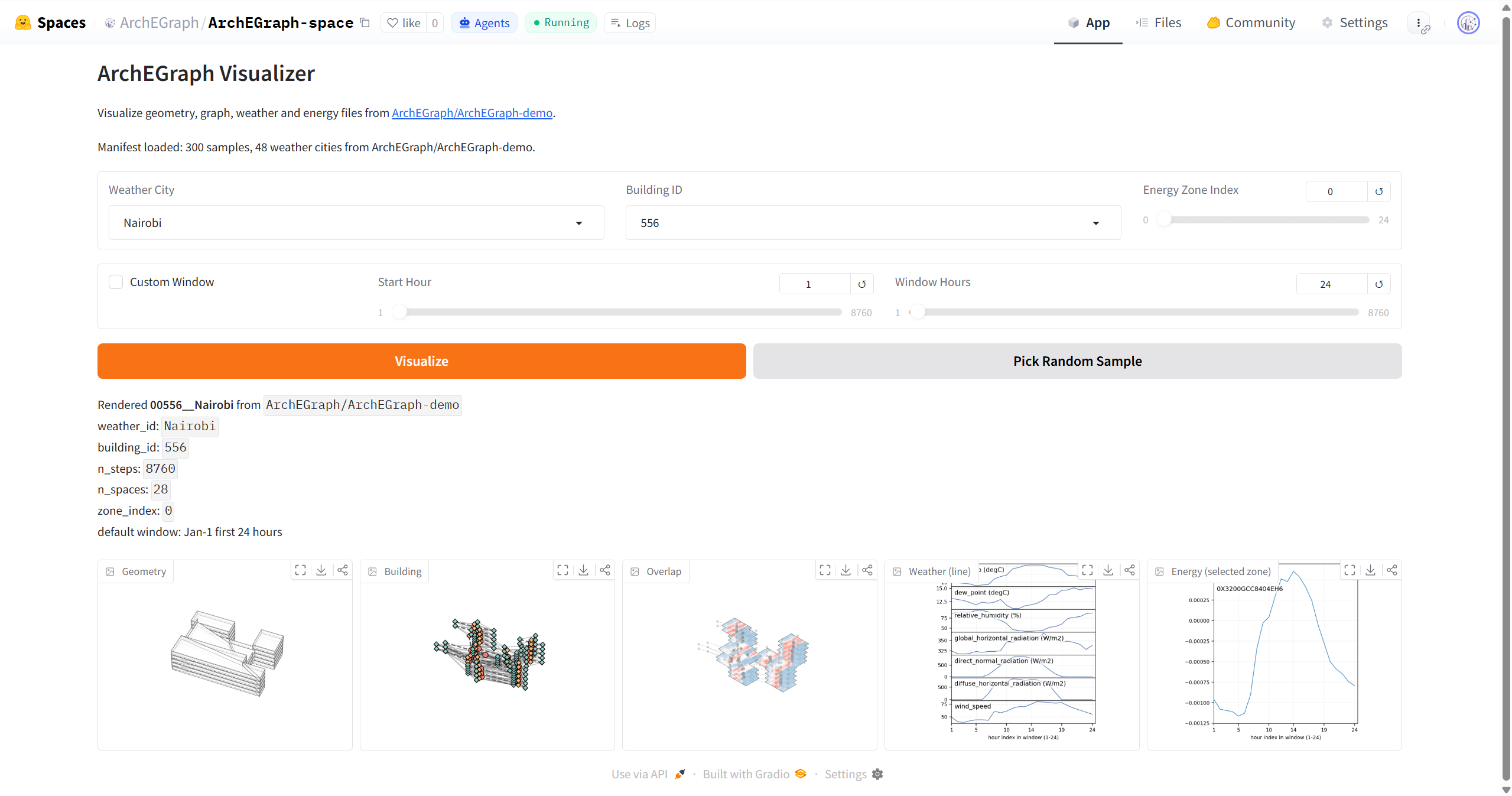}
    }
    \caption{Screenshot of the visualization platform.}
    \label{fig:visualize}
\end{figure}

We provide an interactive online visualization platform for the demo dataset via Hugging Face Spaces: 
\href{https://huggingface.co/spaces/ArchEGraph/ArchEGraph}{ArchEGraph Space}, as shown in Figure~\ref{fig:visualize}.

The platform enables browser-based exploration of the \textbf{ArchEGraph-demo} dataset without local setup. It supports visualization of (i) building geometry and spatial graph structures, (ii) zone-level energy time series, and (iii) coupled weather–energy relationships. Users can navigate across different samples to inspect variations in building–climate configurations. This interface serves as a lightweight tool for qualitative analysis, helping users understand spatial–temporal patterns and verify model behavior on the dataset.

\subsection{Batch visualization of dataset cases}

\begin{figure}[htbp]
    \centering
    \includegraphics[width=\linewidth]{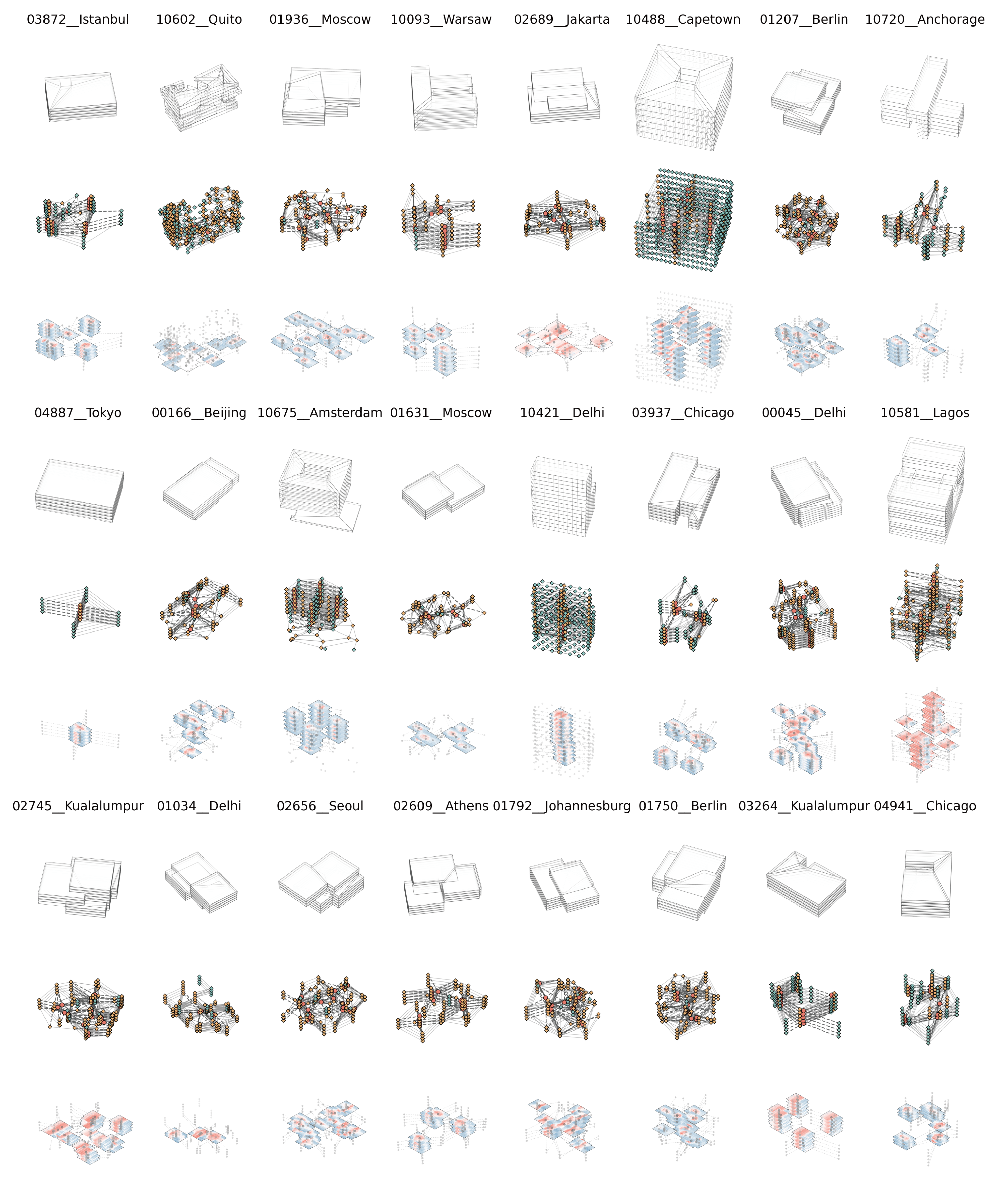}
    \caption{Batch visualization of aligned samples from ArchEGraph across multiple cities. For each case, the triplet shows geometry mesh (top), corresponding heterogeneous graph representation (middle), and zone-level load-map strip (bottom), highlighting cross-case diversity in form, topology, and physical response patterns.}
    \label{fig:batc_visualize}
\end{figure}

The diversity and consistency of ArchEGraph samples are illustrated through a batch-level visualization (shown in Figure~\ref{fig:batc_visualize}). By aligning multiple cases under a unified representation pipeline, this figure provides an intuitive comparison across geometric configurations, topological structures, and their corresponding physical responses. It highlights how variations in architectural form and spatial organization are reflected in both graph structure and simulated load patterns, offering a qualitative overview of the dataset’s richness and cross-case variability.

\section{Broader impacts}
\label{si:impact}
This work may contribute positively to sustainable building design and climate-aware urban development by enabling faster and more scalable building energy analysis. Improved surrogate models for building performance simulation could help architects and engineers evaluate energy efficiency earlier in the design process, potentially reducing operational energy consumption and carbon emissions in the built environment. The dataset may also support broader research in scientific machine learning, graph representation learning, and AI-assisted engineering design.

Potential negative impacts should also be considered. Increased reliance on automated prediction systems in architectural workflows could reduce transparency in decision-making if surrogate models are used without proper physical validation or domain expertise. In addition, AI-driven optimization tools may disproportionately benefit regions or organizations with greater computational and technical resources, potentially widening technological gaps in the architecture, engineering, and construction industries. Finally, although the dataset itself does not contain personal information, future integration of similar systems with real building operation data may raise concerns related to privacy, infrastructure security, or proprietary design information.

\clearpage

\end{document}